\documentclass{clv2-draft}  
\usepackage{epsfig}
\usepackage{plain}
\usepackage{graphicx}
\usepackage[usenames,dvipsnames]{color}
\usepackage{ulem,multirow}
\usepackage{hyperref,url}
\usepackage{amsmath,amssymb,bm}
\usepackage{subfigure}
\usepackage{caption}
\usepackage{lingmacros} 
\usepackage{xspace}
\usepackage{rotating}
\usepackage{setspace}
\usepackage{booktabs}
\usepackage[table]{xcolor}
\usepackage{bold-extra}
\usepackage{arabtex}
\usepackage[activate=false]{microtype}
\usepackage{ragged2e}
{\newdimen\origiwspc%
\newdimen\origiwstr%
\origiwspc=\fontdimen2\font
\origiwstr=\fontdimen2\font
\fontdimen2\font=0.5ex
\fontdimen3\font=0.7em
}%
{\fontdimen2\font=\origiwspc
\fontdimen3\font=\origiwstr
}

\newcommand \COMMENT[1]{}
\newcommand{\TOCHECK}[1]{\textcolor{blue}{#1}}

\newcommand{\MAYBE}[1]{}					

\newcommand{\gap}{\vspace{2mm}}

\newcommand{\eg}{e.\,g.\xspace}
\newcommand{\ie}{i.\,e.\xspace}

\newcommand{\oline}[1]{$\overline{\mbox{#1}}$}

\newcommand\customspacestyle[1]{\SetTracking{encoding=*}{#1}\lsstyle}
\newcommand\normalspacestyle{\SetTracking{encoding=*}{0}\lsstyle}

\DeclareMathSymbol{\mlq}{\mathord}{operators}{``}
\DeclareMathSymbol{\mrq}{\mathord}{operators}{`'}

\renewcommand{\b}{{\bf b}}             
\newcommand{\f}{{\bf f}}               
\newcommand{\e}{{\bf e}}               

\title{A Survey of Word Reordering in Statistical Machine Translation: Computational Models and Language Phenomena}
\author{Arianna Bisazza%
\thanks{Informatics Institute, University of Amsterdam, Science Park 904, 1098 XH Amsterdam, The Netherlands. E-mail: \texttt{a.bisazza@uva.nl} 
}}
\affil{University of Amsterdam}

\author{Marcello Federico%
\thanks{Fondazione Bruno Kessler, Via Sommarive 18, 38123 Povo, Trento, Italy. 
E-mail: \texttt{federico@fbk.eu} 
}}
\affil{Fondazione Bruno Kessler}

\runningtitle{A Survey of Word Reordering in Statistical Machine Translation}
\runningauthor{Bisazza and Federico}

\begin{document}


\setarab 
\arabtrue 
\novocalize

\abovedisplayskip=.9\abovedisplayskip
\abovedisplayshortskip=.9\abovedisplayshortskip
\belowdisplayskip=.9\belowdisplayskip
\belowdisplayshortskip=.9\belowdisplayshortskip

\maketitle

\begin{abstract}


Word reordering is one of the most difficult aspects of statistical machine translation (SMT), and an important factor of its quality and efficiency. 
Despite the vast amount of research published to date, the interest of the community in this problem has not decreased,
and no single method appears to be strongly dominant across language pairs.
Instead, the choice of the optimal approach for a new translation task still seems to be mostly driven by empirical trials.

To orientate the reader in this vast and complex research area,
we present a comprehensive survey of word reordering viewed
as a statistical modeling challenge and as a natural language phenomenon.  
The survey describes in detail how word reordering is modeled within different string-based and tree-based SMT frameworks
and as a stand-alone task, including systematic overviews of the literature in advanced reordering modeling.

We then question why some approaches are more successful than others in different language pairs.
We argue that, besides measuring the \textit{amount} of reordering, it is important to understand which \textit{kinds} of reordering  occur in a given language pair.
To this end, we conduct a qualitative analysis of word reordering phenomena in a diverse sample of language pairs, based on a large collection of linguistic knowledge.
Empirical results in the SMT literature are shown to support 
the hypothesis that a few linguistic facts can be very useful to anticipate the reordering characteristics 
of a language pair and to select the SMT framework that best suits them.

\end{abstract}


\section{Introduction}

Statistical machine translation (SMT) is a data-driven approach to the translation of text from a natural language into another.
Emerged in the 1990s and matured in the 2000's to become widespread today, 
the core SMT methods \cite{Brown:90a,Brown:93,Berger:96:pat,Koehn:03} 
learn direct correspondences between source and target language from collections of translated sentences, 
without the need of abstract linguistic representations.
The main advantages of SMT are versatility and cost-effectiveness:
in principle, the same modeling framework can be applied to any pair of languages with minimal engineering effort, given sufficient amount of translation data.
However, experience in a diverse range of language pairs has revealed that
this form of modeling is highly sensitive to structural differences between source and target language, particularly at the level of word order.

Indeed, natural languages vary greatly in how they arrange sentence components,
and translating words in the correct order is essential to preserve meaning across languages.
In English, for instance, the role of different predicate arguments is determined precisely by their relative position within the sentence.
Consider the translation example in Figure~\ref{fig:running-example}:
Looking at the English glosses of the Arabic sentence, one can see that corresponding words in the two languages are placed in overall similar orders with the notable exception of the verb (\textit{jdd/renewed}), which occurs at the beginning of the Arabic sentence but in the middle of the English one --- more specifically, between the subject and the object. 
To reach the correct English order, three other reorderings are required between pairs of adjacent Arabic words: (\textit{AlEAhl/the-monarch}, \textit{Almgrby/the-Moroccan}), (\textit{dEm/support}, \textit{-h/his}) and (\textit{Alr\}ys/the-president}, \textit{Alfrnsy/the-French}).
This example suggests a simple division of reordering patterns into long-range, or global, and short-range, or local.
However other language pairs display more complex, hierarchical patterns.

\renewcommand{\arraystretch}{1.5}
\begin{figure*}[t]
\begin{footnotesize} \sf
\begin{tabular}{@{\ }c@{\ \ \ \ } p{.93\textwidth}@{\ }}
\hline
\vspace{1mm}
 & \multicolumn{1}{r}{ \small \< jdad Al`Ahl Alm.grby Almlk m.hmd AlsAds d`m --h l-- m^srw` Alr'iys Alfrnsy > } \\
\sc \ src & \scriptsize \sf \ \ \ \it verb \hspace{28mm}  subject \hspace{30mm}  object \hspace{18mm} complement \\
\vspace{-1mm}
& \ \textbf{\oline{\ \ jdd\ \ }} \ \ \ \oline{\ \ AlEAhl \ \ \ \ \ Almgrby \ \ \ \ Almlk \ \ mHmd \ \ AlsAds} \ \ \oline{\ dEm \ \ -h } \ \ \ \oline{ l- \ m\$rwE \ \ \ Alr\}ys \ \ \ \ \ \ Alfrnsy } \\
\vspace{1mm}
& \scriptsize \sf \ \it renewed \ \ the-monarch \ the-Moroccan \ the-King \ Mohamed \ the-sixth \ \ \ support \ his \ \ \ \ \ to \ \ \ project \ \ the-president \ the-French \\
\hline
\vspace{0.5mm}
\sc ref & \ The Moroccan monarch King Mohamed VI \textbf{renewed} his support to the project of the French President \\
\hline
\end{tabular}
\end{footnotesize}
\caption{Arabic source sentence (right-to-left) and English reference translation, taken from the NIST-MT09 benchmark. The Arabic sentence is morphologically segmented by AMIRA (Diab, Hacioglu, and Jurafsky 2004) according to the Arabic Treebank scheme, and provided with Buckwalter transliteration (left-to-right) and English glosses.}
\label{fig:running-example}
\end{figure*}
\renewcommand{\arraystretch}{1}

\nocite{Diab:04}

Word reordering phenomena are naturally handled by human translators\footnote{Nevertheless learning and understanding a new language has been shown to be more difficult when the new language is structurally distant from one's native language \cite{Corder:79}.} but are a major source of complexity for SMT.
In very general terms, 
the task of SMT consists of: breaking the input sentence into smaller units,
selecting an optimal translation for each unit and placing them in the correct order.
Searching for the overall best translation throughout the space of all possible reorderings is, however, computationally intractable \cite{Knight:99}.
This crucial fact has motivated an impressive amount of research around two inter-related questions: namely,
how to effectively restrict the set of allowed word permutations? and
how to detect the best permutation among them?

Existing solutions to these problems range from heuristic constraints, based on word-to-word distances and completely agnostic about the sentence content, to linguistically motivated SMT frameworks where the entire translation process is guided by syntactic structure.
The research in word reordering has advanced together with the core SMT research and has sometimes directed it,
being one of the main motivation for the development of tree-based SMT. 
At the same time, the variety of word orders existing in the world languages
has pressed the SMT community to admit the importance of language-specific knowledge
and to reassess its ambitions towards a universal translation algorithm.

According to the Machine Translation Archive,
a scientific interest in this specific subproblem of MT started around 2006 and kept growing at a rapid pace. 
In 2014, the research papers mainly dedicated to reordering accounted for no less than 10\% of all SMT papers.\footnote{Peer-reviewed conferences, workshops and journal papers listed by the Machine Translation Archive: \url{http://www.mt-archive.info/srch/subjects.htm}}
 Despite the abundant research, word order differences remain among the most important factors of performance in modern SMT systems, and new approaches to reordering are still proposed every year. 
\COMMENT{
Figure~\ref{fig:reo-papers} shows the percentage of SMT research papers mainly dedicated to reordering, according to the Machine Translation Archive.\footnote{\url{http://www.mt-archive.info/}}
The data suggests that a scientific interest in this specific subproblem of MT started around 2006 and continued to grow at a rapid pace until 2010. Still no decreasing trend was observed by the end of 2013.
\begin{figure}[h]
\centering
\includegraphics[width=.54\textwidth]{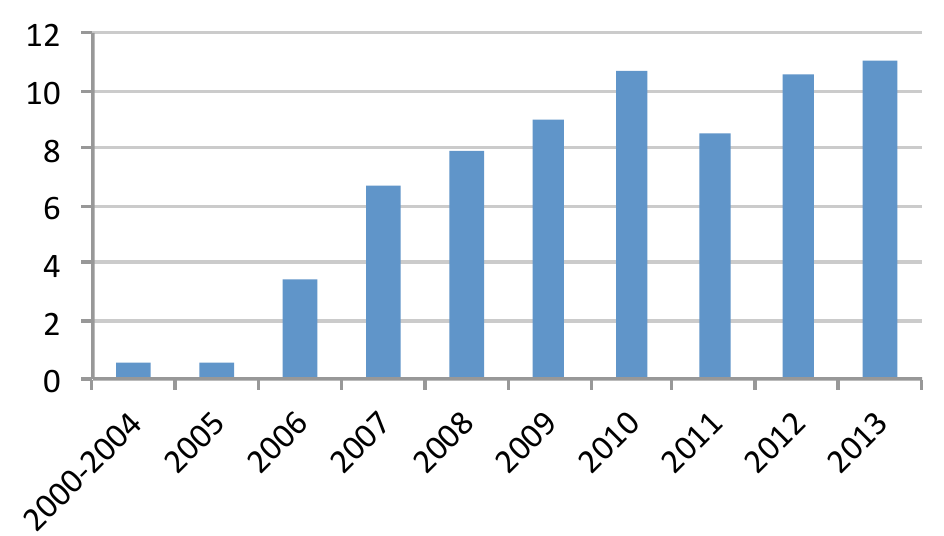}
\caption{\label{fig:reo-papers} Percentage of SMT research papers \TOCHECK{dedicated mainly} to reordering
(peer-reviewed conferences, workshops and journals). Source: Machine Translation Archive.}
\end{figure}
}

To orientate the reader in this complex and productive research area,
we present a comprehensive survey of word reordering viewed
as a statistical modeling challenge and as a natural language phenomenon.
Our survey notably differs from previous \cite{Costa-Jussa:09} in that we do not only review the existing approaches to word reordering in SMT, but we also question why some approaches are more successful than others in different language pairs.
In particular, we argue that understanding the complexity of reordering in a given language pair is key to selecting the right SMT models and to improving them.

The survey is organized as follows:
Section \ref{sect:methods} explains how the word reordering problem is treated within different string-based and tree-based SMT frameworks, as well as a stand-alone task (\ie pre- and post-ordering).
The literature in advanced reordering modeling is extensively reviewed, with a major focus on recent work.
Section \ref{sect:evaluation} describes the challenges of automatically assessing word reordering accuracy in SMT outputs.
Section \ref{sect:phenomena} presents a qualitative analysis of word reordering across language pairs.
In particular, detailed word order profiles are provided for a sample of seven widely spoken languages
representing structural and geographical diversity:
namely English, German, French, Arabic, Turkish, Japanese and Chinese.
The same section reviews empirical results from the SMT literature showing that the proposed word order profiles 
are useful to anticipate the reordering characteristics of a language pair and to select the SMT framework that best suits them.
The survey ends with a discussion of the strengths and weaknesses of the major approaches to reordering in SMT.

\section{Approaches to Word Reordering in Statistical Machine Translation}
\label{sect:methods}

A first important distinction has to be made between word reordering performed as part of the decoding process (Sections~\ref{sec:sota-PSMT} to \ref{sec:sota-tree-smt}) 
and word reordering performed \textit{before} or \textit{after} it as a monolingual task decoupled from the bilingual translation task (Section~\ref{sec:sota-preproc}).

Within the former, we further distinguish between string-based (sequential) approaches
and tree-based (structural) approaches.
\textbf{String-based SMT} (Sections~\ref{sec:sota-PSMT} and \ref{sec:sota-ngram}) treats translation as a \textit{sequential} task: the target sentence is built from left to right while the input units are visited in different orders and no dependencies other than word adjacency are considered.
Subsequently, problem decomposition is applied to the target \textit{string}:
an optimal translation is sought for each prefix of the target translation, from the shortest to the longest. 
\textbf{Tree-based SMT} (Section~\ref{sec:sota-tree-smt}) posits the existence of a \textit{tree structure} 
to explain translation as a hierarchical process and to capture dependencies among non-adjacent text units.
Problem decomposition is therefore based on this structure: an optimal translation is sought for each word span corresponding to a node in the tree, from the leaves up to the root.
Whereas string-based SMT has to search over all input permutations that do not violate some general reordering constraints,
tree-based SMT considers only those permutations that result from transforming a given tree representing the input sentence (as for example permuting each node's children).  

Moreover, we should note the difference between \textbf{syntax-based SMT} approaches that utilize trees produced by monolingual parsers trained on syntactic treebanks,
and \textbf{data-driven tree-based SMT} approaches that extract bilingual translation grammars directly from pairs of source and target sentences. 
In the former, word reordering is constrained by a given syntactic parse tree of the input sentence or by the grammar of the target language (or both),
whereas, in the latter, tree structure 
captures hierarchical reordering patterns that may or may not correspond to syntactically motivated rules.

In general, the SMT search (or decoding) process consists in searching for the most probable target (or English) sentence $\e^*$ given a source (or foreign) sentence $\f$ by scoring translation hypotheses
through a linear combination of feature functions:
\begin{equation}\label{eq:phr-prob}
{\e}^* = \arg\max_{\e} \max_{\b} \ \exp \left[ \sum_{r=1}^R \lambda_r \mbox{h}_r(\f,\e,\b) \right]
\end{equation}
where $\b$ is a latent variable representing either a linear or a hierarchical mapping (alignment) between $\f$ and $\e$, $\mbox{h}_r(\e,\f,\b)$ are R arbitrary feature functions and $\lambda_r$ the corresponding feature weights.  
Feature functions try to capture relevant translation adequacy and word reordering aspects from aligned parallel 
data, as well as translation fluency aspects from monolingual target texts.  Moreover, feature functions are 
assumed to be locally decomposable to allow for efficient decoding via dynamic programming. 
Feature weights are tuned discriminatively by directly optimizing translation quality\footnote{Automatic measures of translation quality are discussed in Section~\ref{sect:evaluation}.} on a development set, using parameter tuning techniques such as MERT \cite{Och:03c}, MIRA \cite{Chiang:2008} or PRO \cite{Hopkins:11}.


\subsection{Phrase-based SMT}
\label{sec:sota-PSMT}
Phrase-based SMT (PSMT) is the currently dominant approach in string-based SMT.
PSMT ruled out the early word-based SMT framework \cite{Brown:90a,Brown:93,Berger:96:pat} 
thanks to two important novelties: the use of multi-word translation units \cite{Och:99,Zens:02,Koehn:03}, 
and the move from a generative to a discriminative modeling framework \cite{Och:02}.

\begin{figure}[t]
\includegraphics[width=\textwidth]{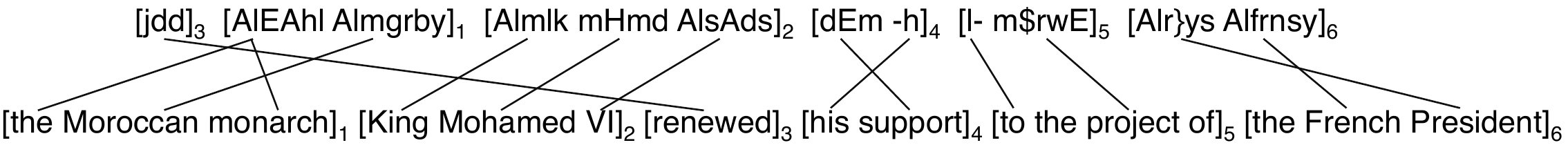} 
\caption{An example of word alignment and phrase segmentation for the sentence pair presented in Figure~\ref{fig:running-example}.
Subscript indices denote the phrase alignment $b_1^I $. Note that other phrase segmentations are possible given the same word alignment.}
\label{fig:running-example-phr}
\end{figure}

The search process (\ref{eq:phr-prob}) in PSMT is guided by the target string $\e$, built from left to right, and the alignment variable $\b$ which embeds both segmentation and reordering of the source phrases. 
This is defined as: \begin{equation}\label{eq:phr-alignment}
\b = b_1^I = ((J_1,K_1),(J_2,K_2),\ldots,(J_I,K_I))
\end{equation}
such that $K_1,\ldots,K_I$ are consecutive intervals partitioning the target word positions,
and $J_1,\ldots,J_I$ are corresponding but not necessarily consecutive intervals partitioning 
the source word positions.  
A phrase segmentation for our running example is shown in Figure~\ref{fig:running-example-phr}.

The use of phrases mainly results in a better handling of ambiguous words and many-to-many word equivalences,
but it also makes it possible to capture a considerable amount of local reordering phenomena within a translation unit (\textit{intra}-phrase reordering).
With reference to our running example (Figure~\ref{fig:running-example}),
a PSMT model may handle the local reorderings as single phrase pairs --- \textit{[AlEahl Almgrby]-[The Moroccan monarch]} etc. --- if these were observed in the training data.
On the contrary, 
it is unlikely that a single long phrase spanning from \textit{jdd} to \textit{AlsAds} was observed,
therefore the long-range reordering of the verb could be handled by \textit{inter}-phrase reordering.

State-of-the art PSMT systems typically include the following core feature functions:
phrase- and word-level translation models;
target n-gram language model; 
distortion penalty;
%
plus additional components that model specific translation aspects. 
%
%
Assuming a one-to-one correspondence between source and target phrases, reordering in PSMT means searching through a set of permutations of the source phrases. Thus, two sub-problems arise: defining the set of permutations in $\b$ allowed during decoding (reordering constraints) and
scoring the allowed permutations (reordering models or feature functions).
We will now discuss each of them in detail.

\gap

\subsubsection{PSMT reordering constraints}
\label{sec:sota-constraints}

Because searching over the space of all possible translations is NP-hard \cite{Knight:99}, 
SMT decoders employ heuristic search algorithms to only explore a promising subset of the search space.
In particular, limiting the set of explorable input permutations is an essential way to reduce decoding complexity.

The reordering constraint originally included in the PSMT framework is called \textbf{distortion limit (DL)}. 
This  consists in allowing the decoder to jump, or skip, at most $k$ words between the last translated source phrase and the next one, i.e.:
\begin{equation}\label{eq:disto-limit}
jump(J_{i-1},J_i) = \left|\displaystyle start(J_i) - end(J_{i-1}) -1 \right| \ \leq \ k
\end{equation}
Setting a low distortion limit means only exploring local reorderings, based on the arguable assumption that languages tend to arrange sentence constituents in similar orders.
Besides being essential for efficiency --- DL allows for linear decoding complexity ---, reordering constraints are also important for translation quality
because the existing SMT models are typically not discriminative enough to guide the search over very large sets of reordering hypotheses. 
However, reordering constraints have also several drawbacks.
For instance the verb reordering in Figure~\ref{fig:running-example-phr} may not be captured by a PSMT system that applies a DL of $k$=5 or less, because jumping back from \textit{AlsAds} to \textit{jdd} corresponds to a skip of 6 positions.
While the distortion limit is a de facto standard in modern PSMT systems,
the first constraining paradigms were formulated earlier for word-based SMT \cite{Berger:96:pat,Zens:03} and are called IBM constraints. 

A different kind of reordering constraint can be derived from the Inversion Transduction Grammars (\textbf{ITG}) \cite{Wu:95,Wu:97}.
\textbf{ITG constraints} only admit permutations that are generated by recursively swapping pairs of adjacent blocks of words.%
\footnote{For a comparative study of the IBM and ITG constraints, we refer the reader to \namecite{Zens:03}.}
In particular, ITG constraints disallow reorderings that generalize the patterns \mbox{(3 1 4 2)} and \mbox{(2 4 1 3)}, which are rarely attested in natural languages \cite{Wu:97}.%
\footnote{Empirical evidence against this was presented by \namecite{Wellington:06}.}
Enforcing ITG constraints in left-to-right PSMT decoding requires the use of a shift-reduce permutation parser \cite{Zens:08:thesis,Feng:COLING10}.
Alternatively, a relaxed version of the ITG constraints (i.e. Baxter permutations) 
may be enforced by simply inspecting the set of covered source positions, as proposed by \namecite{Zens:04} and \namecite{Zens:08:thesis}.
Interestingly, \namecite{Cherry:12} found no consistent benefit from applying either exact or approximate ITG-constraints to a PSMT system that already included a hierarchical phrase orientation model\footnote{The reordering models mentioned herein are explained in detail in the next subsection.} \cite{Galley:08}.

The reordering constraints presented so far are not sensitive to the words being translated nor to their context. 
This results in a very coarse definition of the reordering search space, which is problematic in language pairs with different syntactic structures.
To address this problem, \namecite{Yahyaei:09} propose to decouple local and global reordering by segmenting the input sentence into chunks that can be permuted arbitrarily, but each of which is translated monotonically. 
%
In a related work, \namecite{Yahyaei:10} present a technique to \textbf{dynamically set the DL} during decoding:
they train a discriminative classifier 
to predict the most probable jump length after each input word, 
and use the predicted value as the DL after that position.
Unfortunately, this method appears to generate inconsistent constraints leading to decoding dead-ends.
%
\namecite{Bisazza:13} further develop this idea so that only long reorderings predicted by a specific reordering model are explored by the decoder.
This form of early reordering pruning enables the PSMT system to capture long-range reordering without hurting efficiency and is not affected by the constraint inconsistency problem.

When available, a parse tree of the input may also be used to constrain PSMT reordering following the principle of \textbf{syntactic cohesion} \cite{Fox:02}.
Concretely, the dependency cohesion constraint \cite{Cherry:08} states that, when part of a source subtree is translated, all words under the same subtree must be covered before moving to words outside of it. Integrated in phrase-based decoding as soft constraints  --- \ie by using the number of violations as a feature function --- dependency cohesion and its variants \cite{Cherry:08,Bach:09b} were shown to significantly improve translation quality.
In a related work, \namecite{Feng:12} derive similar cohesion constraints from the semantic role labeling structure of the input sentence.
The divide-and-translate approach of \namecite{Sudoh:10} employs source-side parse trees to segment complex sentences into simple clauses
which are replaced by specific symbols and translated independently.
Then, the target sentence is reconstructed using the placeholders, with the aim of simplifying long-range clause-level reordering.

\subsubsection{PSMT reordering feature functions}
\label{sec:sota-features}

Target language modeling is the primary way to reward promising reorderings during translation.
This happens indirectly, through the scoring of target word n-grams that are generated by translating the source positions in different orders.
However, the fixed-size context of language models used in SMT (typically 4 or 5 words) makes them largely insensitive to global reordering phenomena.
In the last years, a growing interest for language pairs with very different word orders, such as Arabic-English and Chinese-English, has favored the development of new techniques to explicitly model the reordering problem.
Given a source sentence, the search for its optimal reordering is generally decomposed into a sequence of local reordering decisions, as is done for the whole translation process.
Thus, the basic reordering step corresponds to the relative positioning of the word, or phrase, being translated with respect to the word, or phrase, that was previously translated.

The simplest example of reordering feature function is the \textbf{distortion cost} or distortion penalty $jump(J_{i-1},J_i)$, which by convention assigns zero cost to hypotheses that preserve the order of the source phrases (monotonic translations). 
During decoding, the basic implementation of distortion cost penalizes long jumps only when they are performed, leading to the proliferation of hypotheses with gaps (\ie uncovered input positions).
This issue can be addressed by incorporating into the distortion cost an estimate of the cost yet to be incurred \cite{Moore:07}.

\gap

State-of-the-art systems use the distortion cost in combination with more sophisticated reordering models that take into account the identity of the reordered phrases and, optionally, various kinds of contextual information.
A representative selection of such models is summarized in Table~\ref{tab:reomodels}.
To ease the presentation, we have divided the models into four groups according to their problem formulation: phrase orientation models, jump models, source decoding sequence models and operation sequence models.

\renewcommand{\arraystretch}{1.3}
\begin{table}[t]
\setlength{\tabcolsep}{2.5pt}
\centering \footnotesize
\begin{tabular}{| l | l @{\ }|@{\ }l@{\ }|l |l @{\ }| }
\hline
 \multirow{2}{*}{\bf Reordering models}	&  \multirow{2}{*}{\bf References} & \bf Model   & \bf Reordering step & \bf  \multirow{2}{*}{Features} \\
& & \bf type & \bf classification & \\
\hline
\multicolumn{5}{ c }{\textbf{Phrase orientation models (POM):}} \\
\multicolumn{5}{ c }{Example: $P(\textsl{orient=discontinuous-left}\ | \ \textsl{next-phrase-pair=} \textit{[jdd]-[renewed]})$} \\
\hline
  & Tillmann 2004;   & \multirow{4}{*}{gener.}  &   &    \\
lexicalized (hierarchical) & Koehn \& al. 2005;  &  &   &  source/target phrases\\
phrase orientation model	&	Nagata \& al. 2006; & &  &\\
	&	Galley \& Manning 2008 & & monotonic,  swap, &\\
\cline{1-3}\cline{5-5}
phrase orientation   &  \multirow{2}{*}{Zens \& Ney 2006}  &  \multirow{2}{*}{discr.}   &   discontinuous &   \\
maxent classifier 	& & & (left or right) & source/target words \\
\cline{1-3}
sparse phrase &  \multirow{2}{*}{Cherry 2013} & \multirow{2}{*}{discr.} & & or word clusters \\
orientation features &  & & & \\
\hline
\multicolumn{5}{ c }{\textbf{Jump models (JM):}} \\
\multicolumn{5}{ c }{Example: $P(\textsl{jump=}-$5$ \ | \ \textsl{from=}\textit{AlsAds}, \textsl{to=}\textit{jdd })$} \\
\hline
inbound/outbound/pairwise  & Al-Onaizan \& Papineni &  \multirow{2}{*}{gener.}   &   \multirow{2}{*}{jump length}   & \multirow{2}{*}{source words}   \\%
lexicalized distortion	&  2006 & & & \\
\hline
inbound/outbound  &  \multirow{2}{*}{Green \& al. 2010} &  \multirow{2}{*}{discr.}    & jump length based  & source words, POS,  \\
length-bin classifier & & & (9 length bins) & position; sent. length \\
\hline
\multicolumn{5}{ c }{\bf Source decoding sequence models (SDSM):} \\
\multicolumn{5}{ c }{Example: $P(\textsl{next-word=}\textit{jdd} \ | \ \textsl{prev-translated-words=}\textit{AlEahil Almlk mHmd AlsAds})$} \\
\hline
\multirow{2}{*}{reordered source n-gram} &  \multirow{2}{*}{Feng \& al. 2010a} &  \multirow{2}{*}{gener.}		   &  \multicolumn{1}{ c |}{\multirow{2}{*}{---}}	  & source words   \\
 & & & & (9-gram context) \\
\hline
 \multirow{3}{*}{source word-after-word} & Bisazza \& Federico 2013; &  \multirow{3}{*}{discr.} & \multicolumn{1}{ c |}{\multirow{3}{*}{---}} & source words, POS; \\
 & Goto \& al. 2013 & & & source context's words  \\
 &  & & & and POS \\
\hline
\multicolumn{5}{ c }{\bf Operation sequence models (OSM):} \\
\multicolumn{5}{ c }{Example: $ P(\ \textsl{next-operation=generate[\textit{jdd,renewed}]}\ | \ \textsl{prev-operations=generate[\textit{AlsAds,VI}] \ jumpBack[1]}\ )$} \\
\hline
translation/reordering  & Durrani \& al. 2011; &  \multirow{3}{*}{gener.} & insertGap, & source/target words, \\ 
operation n-gram    & Durrani \& al. 2013; & & jumpBack, & POS or word clusters; \\
    & Durrani \& al. 2014 & & jumpForward & prev. $n$ --1 operations \\
\specialrule{1.5pt}{1pt}{1pt}
\end{tabular}
\caption{\label{tab:reomodels} An overview of state-of-the-art reordering models for PSMT. Model type indicates whether a model is trained in a generative or discriminative way.
All examples refer to the sentence pair shown in Figure~\ref{fig:running-example-phr}.}
\end{table}
\renewcommand{\arraystretch}{1.1}


\textbf{Phrase orientation models (POM)} \cite{Tillmann:04,Koehn:05a,Nagata:06,Zens:06a,Li:14}, simply known as lexicalized reordering models, 
predict whether the next translated source span should be immediately to the right (monotone), immediately to the left (swap) or anywhere else (discontinuous) relatively to the last translated one.\footnote{Some phrase orientation models further distinguish between discontinuous left and discontinuous right.}
For example, in Figure~\ref{fig:running-example-phr}, the phrase pair \textit{[Almlk mHmd AlsAds]-[King Mohamed VI]}  has monotone orientation 
whereas \textit{[jdd]-[renewed]} has discontinuous left orientation with respect to the previously translated phrase. 
%
Because of their simple reordering step classification, POM can be conditioned on very fine-grained information, such as the whole phrase pair, without suffering too much from data sparseness.
However, since POM ignore the distance between consecutively translated phrases, they cannot properly handle long-range reordering phenomena and are typically used with a low distortion limit.

\textbf{Jump models (JM)} \cite{Al-Onaizan:06,Green:10}
predict the direction and length of the jump that is performed between consecutively translated words or phrases,
with the goal of better handling long-range reordering. 
Due to data sparseness, JM work best when trained in a discriminative fashion using a variety of binary features (such as the last translated word, its POS tag and relative position in the sentence) and when length bins are used instead of the exact jump length \cite{Green:10}.
A drawback of JM is that they typically over-penalize long jumps because they are more rarely observed than short jumps. 

\textbf{Source decoding sequence models (SDSM)} address the said issue by directly modeling the reordered sequence of input words, as opposed to the reordering operations that generated it.
This in turn can be done in several ways, such as:
training n-gram models on target-like reordered source sentences and using them to score the sequence of input words visited by the decoder \cite{Feng:10};
tagging the whole input sentence with symbols denoting how each word should be reordered with respect to its left and right context, then rewarding the decoding paths that most agree with the tag sequence \cite{Feng:13}; and finally, predicting which input position
is likely to be translated right after a given input position by means
of a maximum entropy model using word and context features  \cite{Bisazza:13,Goto:13a}.

\textbf{Operation sequence models (OSM)} \cite{Durrani:11} are n-gram models that include lexical translation operations and reordering operations (\textsl{insertGap}, \textsl{jumpBack} or \textsl{jumpForward}) in a single generative story, thereby combining elements from the previous three model families.
An operation sequence example is provided in the lower part of Table~\ref{tab:reomodels}.
OSM are closely related to n-gram based SMT models (see next section) 
but have been successfully applied as feature functions to PSMT \cite{Durrani:13b}.
To overcome data sparseness, OSM can be successfully applied to POS-tags and unsupervised word clusters \cite{Durrani:14}.

\gap


SDSM and OSM have been proven optimal for language pairs where high distortion limits are required to capture  long-range reordering phenomena \cite{Durrani:11,Bisazza:13:WMT,Goto:13a}.
Nevertheless POM remains the most widely used type of phrase-based reordering model and is considered a necessary component of PSMT baselines in any language pair.
%
In particular, two variants of POM deserve further attention because of their notable effect on translation quality: 
hierarchical POM and sparse phrase orientation features.

\begin{figure}[t]\centering
\includegraphics[width=.76\textwidth]{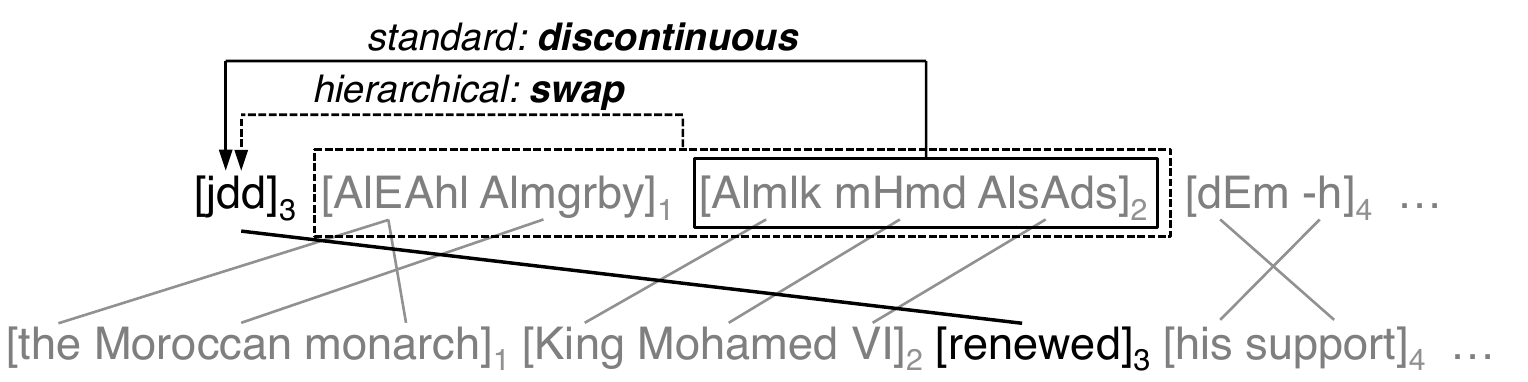}
\caption{\label{fig:running-example-orient} Phrase orientation example for the phrase pair \textit{[jdd]-[renewed]}: 
the standard model detects a discontinuous orientation with respect to the last translated phrase (2)
whereas the hierarchical model detects a swap with respect to the block of phrases (1-2).}
\end{figure}

Hierarchical phrase orientation models, or simply \textbf{hierarchical reordering models} (HRM) \cite{Galley:08} 
improve the way in which the orientation of a new phrase pair is determined: 
already translated adjacent blocks are merged together to form longer phrases around the current one. For instance in Figure~\ref{fig:running-example-orient}, HRM merges phrases $1$ and $2$ into a large phrase pair \textit{[AlEahl ... AlsAds]-[The ... VI]} and consequently assigns a swap, instead of discontinuous orientation, to \textit{[jdd]-[renewed]}.
As a result, orientation assignments become more consistent across hypotheses with different phrase segmentations.

Rather than training a reordering model by relative frequency or maximum entropy and using its score as one dense feature function,
\namecite{Cherry:13} introduces 
\textbf{sparse phrase orientation features} that are directly added to the model score during decoding (cf. equation~(\ref{eq:phr-prob})) and optimized jointly with all other SMT feature weights.
Effective sparse reordering features can be obtained by simply coupling a phrase pair's orientation with the first or last word (or word class) of its source and target side \cite{Cherry:13}, or even with the whole phrase pair identity \cite{Auli:14}.

\subsection{N-gram based SMT}
\label{sec:sota-ngram}

N-gram based SMT \cite{Casacuberta:04a,Marino:06} is a string-based alternative to PSMT. 
In this framework, smoothed n-gram models are learnt over sequences of minimal translation units (called \textit{tuples}), which, like phrase pairs, are pairs of word sequences extracted from word-aligned parallel sentences.
Tuples, however, are typically shorter than phrase pairs and are extracted from a unique, \textit{monotonic} segmentation of the sentence pair.
Thus, the problem of spurious phrase segmentation is avoided but non-local reordering becomes an issue.
For instance, in Figure~\ref{fig:running-example-phr}, a monotonic phrase segmentation could be achieved only by treating the large block \textit{[jdd ... AlsAds]-[The ... renewed]} as a single tuple.
Reordering is then addressed by `tuple unfolding' \cite{Crego:05b}: 
that is, during training the source words of each translation unit are rearranged in a target-like order so that more, shorter tuples can be extracted.
At test time, input sentences have to be \mbox{\textit{pre-ordered}} for translation.
To this end, \namecite{Crego:06b} propose to precompute a number of likely permutations of the input using POS-based rewrite rules learned during tuple unfolding. The reorderings thus obtained are
used to extend the search graph of a monotonic decoder.%
\footnote{More pre-ordering techniques will be discussed in Section~\ref{sec:sota-preproc}.}
Reordering is often considered as a shortcoming of n-gram based SMT as reordering decisions are largely decoupled from decoding and mostly based on source-side information.

\COMMENT{
\gap
\textbf{Discontinuous phrase-based SMT} \cite{Simard:05,Galley:10}
aims at combining the efficiency of string-based decoding with the higher generalization capabilities of discontinuous phrases used in hierarchical SMT (see Section~\ref{sec:sota-tree-smt}).
In this approach, translation units can contain gaps of either fixed \cite{Simard:05} or variable \cite{Galley:10} size,
making it possible to learn more general non-hierarchical reordering patterns.
}

\subsection{Tree-based SMT} 
\label{sec:sota-tree-smt}

The SMT frameworks discussed so far learn direct correspondences between source and target words or phrases,
treating reordering as a sequential process.
This flat representation is fairly successful for some language pairs however,
in others, reordering is more naturally described as a hierarchical process where 
small, locally reordered blocks become the elements of recursively larger reordered blocks.
Concretely, in our running example (Figure~\ref{fig:running-example-phr}), a hierarchical or tree-based approach would make it possible to:
first translate and reorder small blocks such as \textit{[AlEahl Almgrby]} and \textit{[Almlk mHmd AlsAds]},
then merge them to compose a larger block that gets reordered \textit{as a whole} with respect to the verb  \textit{jdd}, and so forth.
The degree of generalization at each level would then depend on how blocks are represented: e.g. by their lexical content, by a tag denoting the block's syntactic category, or by a generic symbol.

Tree-based approaches are largely inspired by syntactic parsing, but not all in the same way:
some model translation as the transformation of trees produced by monolingual parsers trained on syntactic treebanks (Section~\ref{sect:syntax-smt}),
while others extract a bilingual translation grammar directly from word-aligned parallel text without using any syntactic information (Section~\ref{sect:hiero-smt}). 
Non-syntactic bilingual translation grammars may still be enriched with syntactic information, for instance in the form of soft constraints
(Section~\ref{sect:hiero-syntax-smt}).


All tree-based frameworks crucially differ from PSMT and other string-based frameworks with respect to reordering:
Whereas PSMT considers all input permutations that do not violate general reordering constraints 
and then scores them with separate reordering models,
tree-based systems model reordering jointly with translation
and, during decoding, only (or mostly) explore input permutations that are licensed by the learnt translation model.

Most modern tree-based approaches fall under the general formulation of SMT which scores translation hypotheses by a linear combination of feature functions (see equation~(\ref{eq:phr-prob})), 
with a translation model (or grammar) and a target language model as core features.
Tree-based decoding is usually performed by a chart parsing algorithm with beam search and integrated target language model.
Hence, the target sentence is not produced from left to right as in string-based SMT, but bottom-up according to a tree derivation order.

\COMMENT{
To date, many frameworks have been proposed to model translation via tree-like structures, 
such as \namecite{Wu:96,Yamada:02,Galley:04,Imamura:05,Chiang:05a,Ding:05,Quirk:05,Watanabe:06,Shen:10} \textit{\mbox{inter alia}}.
Moreover, some approaches model translation as the transformation of syntactic parse trees produced by pre-trained monolingual parsers \cite{Yamada:02,Galley:04,Quirk:05}, while others extract a bilingual translation grammar directly from word-aligned parallel data without using syntactic information \cite{Wu:96,Chiang:05a}.
For a comprehensive overview of the syntax-based SMT field we refer the reader to \namecite{Koehn:10}.
}

\subsubsection{Syntax-based SMT} 
\label{sect:syntax-smt}


An important motivation for using syntax in SMT is that reordering among natural languages very often involve the permutation of whole syntactic constituents (see e.g. \namecite{Fox:02}).
For instance, in our running example (Figure~\ref{fig:running-example-phr}), knowing the span of the Arabic subject would be enough to predict the reordering of the verb for translation into English.

Syntax-based SMT encompasses a variety of frameworks that use syntactic annotation either on the source or on the target language, or to both.
So-called \textbf{tree-to-string} methods \cite{Huang:06c,Liu:06:TTS} use a given input sentence parse tree to resrict the application of translation/reordering rules to word spans that coincide with syntactic constituents of specific categories.
For instance, the swap of \textit{Alr\}ys Alfrnsy} may only be dictated by a rule applying to noun phrases composed of a noun and an adjective.
On the other hand, \textbf{string-to-tree} methods \cite{Yamada:02a,Galley:04,Marcu:06,Shen:10} employ syntax as a way to restrict translation hypotheses to well-formed target language sentences --- ruling out, for instance, a translation that fails to reorder the translated verb \textit{renewed} with respect to its subject.
%
Using syntax on both source and target sides (\textbf{tree-to-tree}) 
\cite{Imamura:05,Ding:05,Smith:06,Watanabe:06,Zhang:08}
has proven rather difficult in practice due to the complexity of aligning potentially very different tree topologies and to the large size of the resulting translation grammars. Moreover, the need for high-quality parsers in both language sides seriously limits the applicability of this approach.

Syntax-based SMT approaches also differ in the formalism they use to represent the trees.
Those based on phrase structure (constituency) grammars typically comply to the principle that each translation/reordering rule should match a complete constituent, while 
those based on dependency grammars opt for a more flexible use of structure.
%
%
For example, in \textbf{string-to-dependency SMT} \cite{Shen:10} rules can correspond to partial constituents but must be either a single rooted tree, with each child being a complete sub-tree, or a sequence of siblings, each being a complete sub-tree.
Partial dependency rules are then combined during decoding, which means that not all reordering decisions are governed by the translation model.


\begin{figure}[t]
\centering
\includegraphics[width=.85\textwidth]{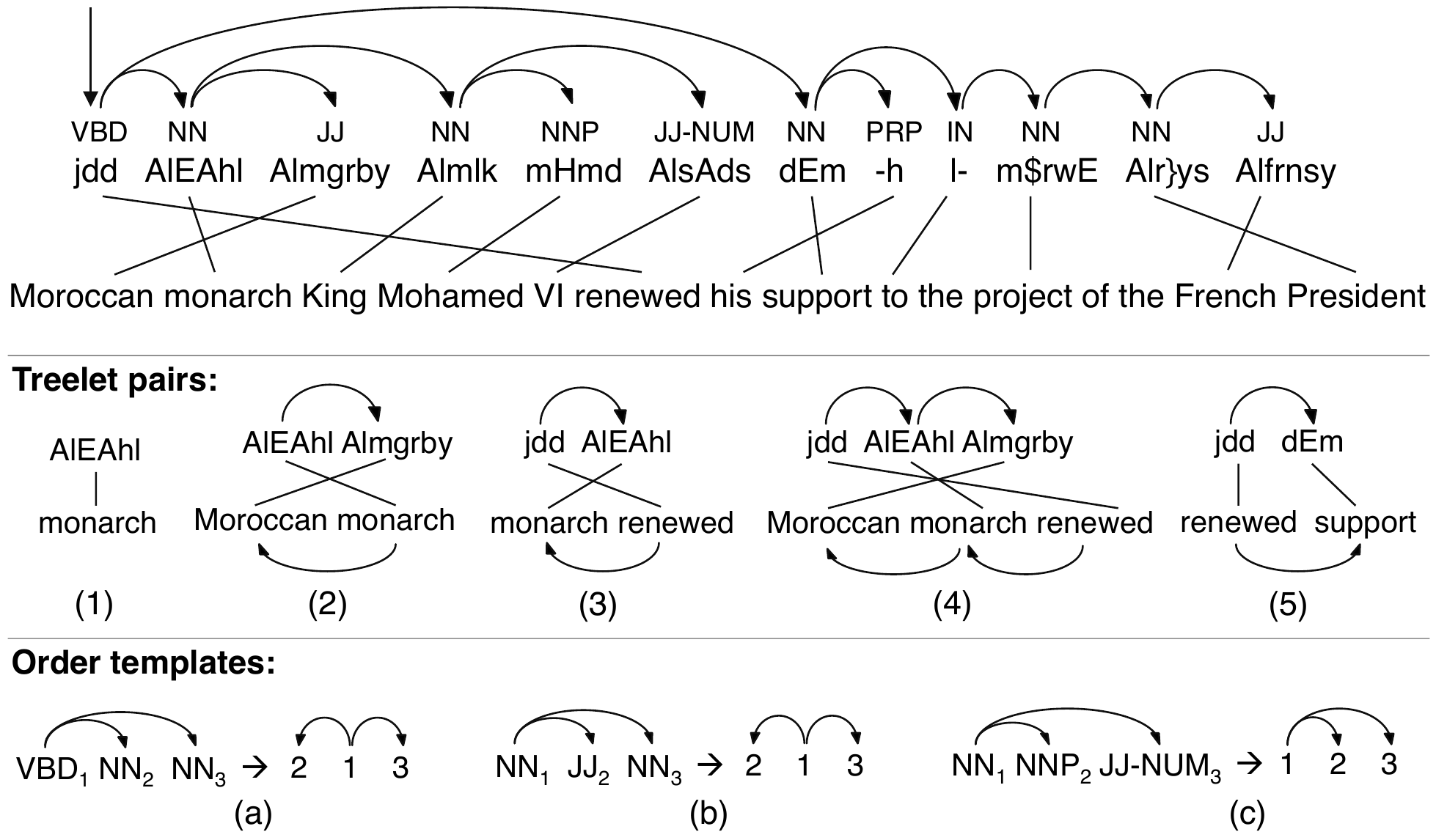}
\caption{\label{fig:running-example-treelet} Examples of treelet pairs and order templates extracted from a word-aligned sentence pair and its source-side dependency parse tree. The projected tree for the whole target sentence is not shown due to space limitations.}
\end{figure}

An even more flexible use of structure is advocated by the \textbf{treelet-based SMT} framework \cite{Quirk:05}, where translation rules can correspond to any connected subgraph of the dependency tree (i.e. treelet).
As illustrated by Figure~\ref{fig:running-example-treelet}, treelet pairs are extracted from pairs of source dependency parse tree and target-side projected tree. 
Treelets can be seen as phrases that are not limited to sets of adjacent words,
but rather to sets of words that are connected by dependency relations,
which in turn makes it possible to learn non-local reordering patterns.
As reordering decisions are only partially governed by the translation model, treelet-based SMT benefits from additional model components specifically dedicated to reordering.
For example, in Figure~\ref{fig:running-example-treelet}, treelet pair (3) determines the swapping of \textit{jdd} and \textit{AlEAhl} 
but does not specify the ordering of \textit{dEm} which is also a child of \textit{jdd}.   
Hence,  during decoding, all possible reorderings of the unmatched children are considered
and scored by a separate discriminative model predicting the position of a child node (or modifier $m$) relative to its head $h$, 
given lexical, POS and positional features of $m$ and $h$.
%
Reordering modeling is thus largely decoupled from lexical selection, which makes the model very flexible but results in a very large search space and high risk of search errors.
To address this issue, \namecite{Menezes:07} introduce another mechanism to complement treelet reordering: namely, dependency order templates.
An order template is an unlexicalized rule specifying the reordering of a node and all its children based on their POS tags.
For instance, in Figure~\ref{fig:running-example-treelet}, treelet pair (3) may be combined with template (a) to specify the order of the child \textit{dEm}.
For each new test sentence, matching treelet pairs and order templates are combined to construct lexicalized translation rules for that sentence and, finally, decoding is performed with a chart parsing algorithm. 

\gap

We will now discuss SMT frameworks that model translation as a process of parallel parsing of the source and target language via a synchronous grammar.


\subsubsection{Tree-based SMT without syntax}
\label{sect:hiero-smt}

The idea of extracting bilingual translation (i.e. synchronous) grammars directly from word-aligned parallel data originates in early work on Inversion Transduction Grammars (ITG) by \namecite{Wu:96,Wu:97}.
%

\begin{figure}[t]
\hspace{-1mm}
\includegraphics[width=1.01\textwidth]{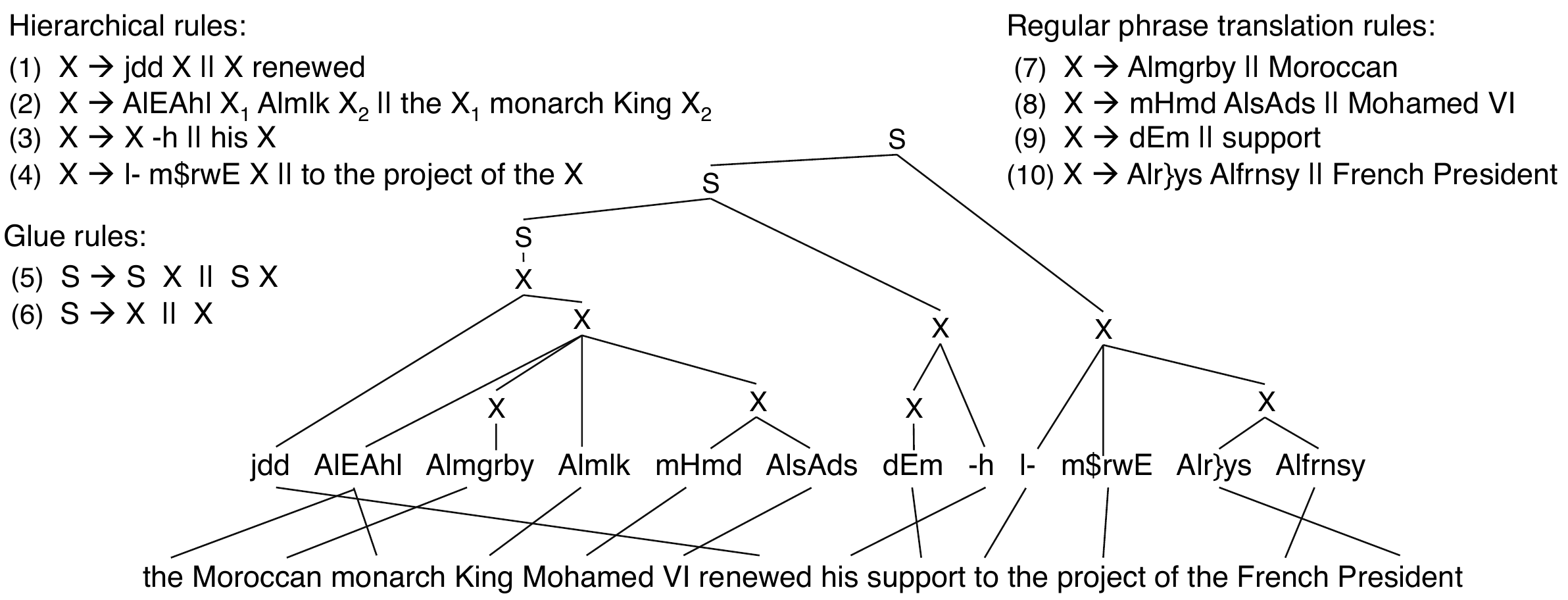}
\caption{\label{fig:running-example-hiero} Possible derivation of a word-aligned sentence pair and  corresponding hierarchical phrase-based translation grammar. 
The target-side tree is not represented due to space limitations.}
\end{figure}

In a more mature approach, \textbf{hierarchical phrase-based SMT (HSMT)} \cite{Chiang:05a}, the translation model is a probabilistic synchronous context-free grammar (SCFG) 
whose rules can correspond to arbitrary (i.e. non syntactically motivated) phrases
labeled by only two generic non-terminal symbols (X or S).
As shown in Figure~\ref{fig:running-example-hiero}, HSMT translation rules can either include:
a mix of terminals and non-terminals capturing reordering patterns and discontinuities (rules 1-4) or
only terminals (7-10) basically corresponding to phrase pairs in string-based PSMT.
Finally, the so-called glue rules (5-6) are always added to the grammar to combine translated blocks in a monotone fashion regardless of their content.
As in PSMT, extracted translation rules may not exceed a certain length and rule scores are obtained using maximum likelihood estimation.
Crucially, swapping adjacent phrases with no lexical evidence ($\textsc{x} \rightarrow \textsc{x}_1 \textsc{x}_2 || \textsc{x}_2 \textsc{x}_1$) is not allowed by standard HSMT grammars,
therefore reordering can only be triggered by at least partially lexicalized translation rules.
This is a major difference with respect to most syntax-based approaches where reordering can be captured by rules containing only labeled non-terminals (e.g. \ $\textsc{s} \rightarrow \textsc{np}\ \textsc{vp}\ ||\ \textsc{vp}\ \textsc{np}$).
This means that, for instance, the reordering pattern learnt by our example HSMT grammar (Figure~\ref{fig:running-example-hiero}, rule 1)
may only be used to reorder the specific verb form \textit{jdd (renewed)} in subsequent test sentences.
Thus, HSMT is likely to work better for languages where the syntactic role of phrases is mostly expressed by separate function words (e.g. Chinese) than for languages where this information is largely conveyed by word inflection (e.g. Russian). 

While hierarchical models are inherently capable of dealing with complex and recursive reordering patterns,
in practice many translation rules are noisy or based on limited context.
To limit search complexity, a constraint is imposed on the maximum number of source words that may be covered by a non-terminal symbol during decoding (\textbf{span constraint}).
This parameter is typically set to 10 or 15 words as wider spans result in prohibitively slow decoding and lower translation quality.
For these reasons, a number of extensions to the original HSMT framework have been proposed with the specific goal of better handling complex reordering phenomena.

\begin{figure}[t]
\hspace{-2mm}
\includegraphics[width=.93\textwidth]{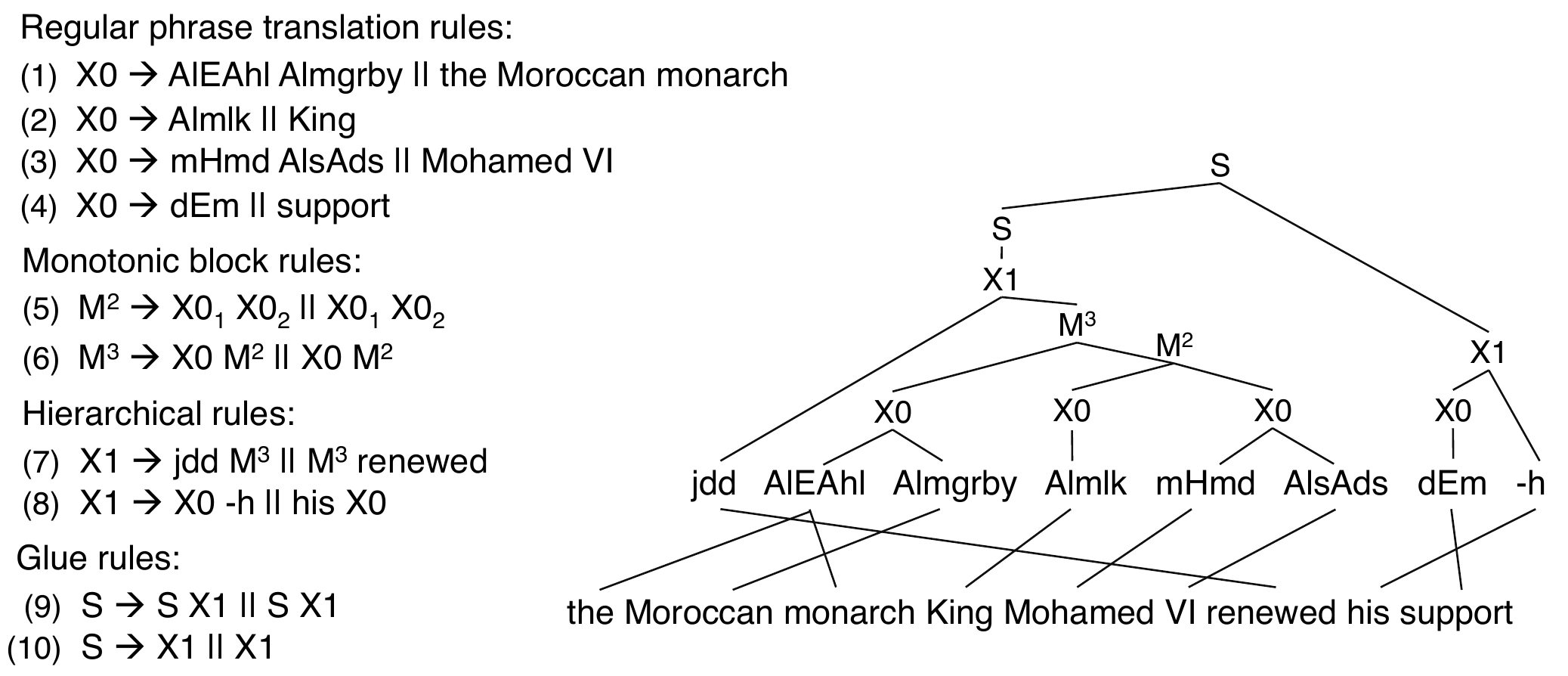}
\vspace{2mm}
\caption{\label{fig:running-example-hiero-s1} Example of shallow-1 HSMT grammar with monotonic \mbox{non-terminals} M$^k$. The target-side tree is not represented due to space limitations.}
\end{figure}

\textbf{Shallow-\textit{n} grammars} \cite{deGispert:10} can be used to refine the reordering space of HSMT according to the reordering characteristics of a specific language pair.
For instance, as shown in Figure~\ref{fig:running-example-hiero-s1}, 
an Arabic-English HSMT grammar is extended with an additional non-terminal symbol X0 
that can only generate fully lexicalized phrases, thereby disallowing recursive nesting of hierarchical rules (shallow-1 grammar).
To account for the movement of large word blocks, 
other new non-terminals M$^k$ allow for the monotonic generation of $k$ non-terminals X0.
While defining a much smaller search space than the original HSMT grammar,
the resulting shallow grammar can capture the long-range reordering of our running example 
even in the likely absence of a rule covering the whole subject span (that is rule~2 in Figure~\ref{fig:running-example-hiero}).

In a related work specifically addressing the issue of long-range reordering, \namecite{Braune:12} 
propose to \textbf{relax the span constraint} only for specific types of hierarchical rules which are more likely to capture long reordering patterns in German-English.
For instance, rules whose source side 
starts with at least one terminal followed by one non-terminal and ends with at least one terminal ($t^{+}\ \textsc{x} \ t^{+}$)
can capture the pattern `finite-auxiliary-verb X participle' (\eg \textit{ist X gestiegen/has increased X}) with very wide X spans.

\namecite{Mylonakis:10} separate the modeling of local reordering (captured by fully lexicalized phrase-pair emission rules) from the modeling of higher-order recursive reordering (captured by ITG-style non-lexicalized binary rules).
Instead of a single non-terminal X, three different \textbf{reordering-based labels} are used according to the reordering pattern in which they participate: X for monotonic rules; XSL and XSR for the first and second symbol, respectively, of swapping rules.
Thus reordering decisions are conditioned on the phrase pair's content, rather than its lexical context as in HSMT.
More fine-grained non-terminals are introduced by \namecite{MailletteDeBuy:14} to also capture the relation of a phrase pair's reordering with respect to the parent phrase that contains it.


Rather than relabeling non-terminals, other work incorporates \textbf{reordering-specific models as additional feature functions}.
\namecite{He:amta10} add to their HSMT grammar the generic phrase swapping rule ($\textsc{x} \rightarrow \textsc{x}_1 \textsc{x}_2 || \textsc{x}_2 \textsc{x}_1$) and use a maximum-entropy classifier to predict whether two neighboring phrases should be swapped or not during decoding.
Rather than conditioning the decision on the whole phrase pair, the classifier employs features extracted from it, such as first and last word (or POS tag) of the source and target side.
A similar model was first developed by \namecite{Xiong:06} for simpler phrase translation models (\ie without discontinuities) based on ITG. 
\namecite{Li:13} use recursive autoencoders \cite{Socher:11} 
to assign vector representations to the neighboring phrases given as input to the ITG classifier,
thereby avoiding manual feature engineering but affecting hypothesis recombination and decoding speed.
\namecite{Nguyen:13} and \namecite{Huck:13} successfully integrate the distortion cost feature function and phrase orientation models initially designed for string-based PSMT into a chart-based HSMT decoder.


Finally, \namecite{Setiawan:07} observe that, in languages like Chinese and English, function words provide important clues on the grammatical relationships among phrases.
Consequently, they introduce a SCFG where \textbf{function words} (approximated by high-frequency words) are the only lexicalized non-terminals guiding phrase reordering.
Based on the same intuition, \namecite{Setiawan:09} augment a HSMT system with a function-word ordering model that predicts, for any pair of translation rules, which one should dominate the other in the hierarchical structure, based on the function words that they contain.%
\footnote{Two other models utilizing function words as the \textit{anchors} of global reordering decisions are proposed in \protect\cite{Setiawan:13:ACL} and \protect\cite{Setiawan:13:EMNLP}. Although integrated in a syntax-based system \protect\cite{Shen:10}, these models are in principle applicable to other SMT frameworks such as HSMT.}

\subsubsection{Tree-based SMT with soft syntactic constraints} 
\label{sect:hiero-syntax-smt}

We have discussed SMT frameworks where the translation model is fully based on the syntactic parse tree of the source or target sentence (Section~\ref{sect:syntax-smt}) 
or where syntax is not used at all (Section~\ref{sect:hiero-smt}).
%
A third line of work bridges between these two by 
exploiting syntactic information in the form of soft constraints
while operating with a synchronous translation grammar extracted from non-parsed parallel data.


%
\namecite{Chiang:05a} first experimented with a feature function rewarding translation rules applied to full syntactic constituents (\textbf{constituent feature}).
While this initial attempt did not appear to improve translation quality,
\namecite{Marton:08} further elaborated the idea and proposed a series of finer-grained features distinguishing among constituent types (\textsc{vp, np}, etc.), eventually leading to better performance.
\namecite{Gao:11} extract two reordering-related feature functions from source dependency parse trees:
(i) The \textbf{dependency orientation model} predicts whether the relative order of a source word and its head should be reversed during translation. This is trained as a maximum-entropy classifier using the words and their dependency relation type as features.
(ii) The \textbf{dependency cohesion penalty} fires whenever a word and its head are translated separately (\ie by different translation rules)
thereby measuring derivation well-formedness.
%
Since long-range reordering tends to happen closer to the root and local reordering closer to the leaves,
a distinction is made between words occurring at different depths of the dependency tree leading to a number of sub-features.
In this way, the tuning process can decide how important or reliable are feature scores coming from different levels of the parse tree.
\namecite{Huang:13} worked instead with constituency parses
and trained a classifier to predict whether the order of any two sibling constituents in the input tree should be reversed or maintained during translation.
The classifier is trained by maximum-entropy using a number of syntactic features
and used during decoding at the word level: that is, each pair of input words inherit the orientation probabilities of the constituents that cover them respectively.


Syntactic annotation has been also used to \textbf{refine non-terminal SCFG labels},
potentially leading to better reordering choices. 
In \cite{Zollmann:06} and \cite{Mylonakis:11}, labels indicate whether a phrase corresponds to a syntactic constituent or to part of it, as well as the constituent type,
relatively to a target or source parse tree, respectively.
Moreover, \namecite{Mylonakis:11} treat the phrase-pair category as a latent variable and let their system learn reordering distributions over multiple labels per span (e.g. generic X or source-syntax based like \textsc{np}, \textsc{vbz+dt}, etc.).
\namecite{Li:wmt12} use source dependency annotation to refine non-terminal symbols with syntactic head information. More specifically, given a hierarchical phrase, its type is obtained by concatenating the POS tags of the exposed heads it contains on the source side, where an exposed head is a word dominated by a word outside the phrase.
Like \namecite{He:amta10}, \namecite{Li:wmt12} also allow adjacent phrases to swap but instead of introducing a separate orientation model, they rely on rule translation probabilities based on the refined non-terminals to guide reordering. 

\subsection{Word Reordering as Pre- (or Post-) Processing}
\label{sec:sota-preproc}

\begin{figure}[t]\centering
\includegraphics[width=.68\textwidth]{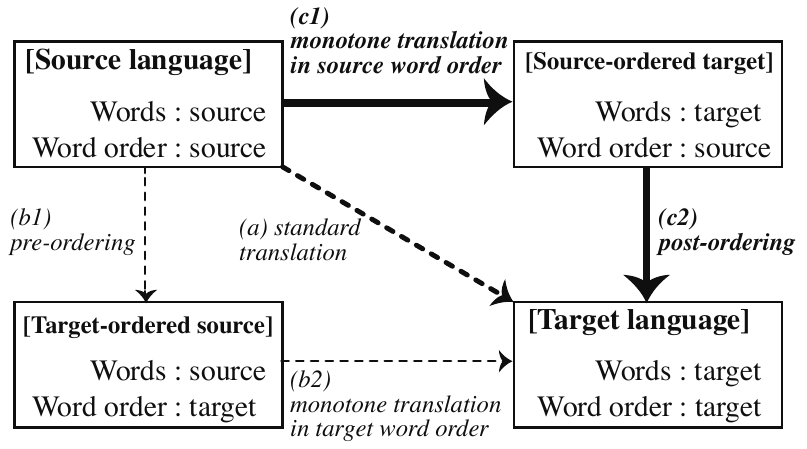} 
\caption{Typical workflows of standard, pre-ordering and post-ordering approaches to SMT. Taken from \protect\namecite{Sudoh:11}.}
\label{fig:sudoh-diagram}
\end{figure}

Given the complexity of solving word reordering during the decoding process,
a productive line of research has focused on decoupling  reordering decisions from  translation decisions.
These approaches aim at arranging words in a target-like order either on the input, \textit{before} translating, or
on the \textit{output}, after translating.
Thus, word reordering is solved as pre- or post-processing (\ie \textit{pre-ordering} or \textit{post-ordering}) in a monolingual fashion and with unconstrained access to the whole sentence context.
Figure~\ref{fig:sudoh-diagram} \cite{Sudoh:11} illustrates the workflows of pre- and post-ordering approaches as opposed to standard SMT.

\subsubsection{Main pre-ordering strategies}

A large number of pre-ordering strategies has been proposed. As a first classification, we divide them into deterministic, non-deterministic and hybrid.
%
\textbf{Deterministic pre-ordering} aims at finding a single optimal permutation of the input sentence, which is then translated monotonically or with a low distortion limit \cite{Niessen:01,Xia:04,Collins:05,Popovic:06p,Costa-jussa:06,Wang:07b,Habash:07,Li:07,Tromble:09,Xu:09,Genzel:10,Isozaki:10b,Yeniterzi:10,Khalilov:11j,Khalilov:11,Visweswariah:11,Gojun:12,Yang:12,Lerner:13,Jehl:14}.\footnote{\namecite{Li:07} experiment with a small number of \textit{n}-best pre-orderings given as alternative inputs to the SMT system.} 
\textbf{Non-deterministic pre-ordering} encodes multiple alternative reorderings into a word lattice and lets a monotonic  (usually n-gram based) decoder choose the best path according to its models \cite{Zens:02,Kanthak:05,Crego:06b,Zhang:07,Rottmann:07,Crego:08,Elming:09,Niehues:09}.
A \textbf{hybrid approach} is adopted by \namecite{Bisazza:2010:WMT,Andreas:11}:
rules are used to generate multiple likely pre-orderings, but only for specific language phenomena that are  responsible for difficult (long-range) reordering patterns. 
The sparse reordering lattices produced by these techniques are then translated by a decoder performing additional phrase-based reordering.
%
In a follow-up work, \namecite{Bisazza:12b} introduce another way to encode multiple pre-orderings of the input: instead of generating a word lattice, 
pre-computed permutations are represented by a {\bf modified distortion matrix} so that lower distortion costs or `shortcuts' are permitted between selected pairs of input positions.


\gap

Pre-ordering methods can also be classified by the kind of pre-ordering rules that they apply:
that is, manually written based on linguistic knowledge, or automatically learned from data. 
We now discuss each of them in detail.

\vspace{-2mm}

\subsubsection{Linguistic knowledge based pre-ordering} In these approaches,
manually written rules determine the transformation of input syntax trees \cite{Collins:05,Wang:07b,Xu:09,Isozaki:10b,Yeniterzi:10,Gojun:12,Andreas:11}
or the permutation of shallow syntactic chunks in a sentence \cite{Hardmeier:2010:WMT,Durgar-El-Kahlout:10,Bisazza:12j}.
In an early example of syntax-based pre-ordering, \namecite{Collins:05} propose a set of six rules aimed at arranging German sentences in English-like order. 
The rules address the position of verbs, verb particles and negation particles, and they are applied to constituency parse trees. 
Following a similar approach, \namecite{Gojun:12} develop a set of rules for the opposite translation direction (English-to-German).
\namecite{Xu:09} instead propose a simple set of dependency-based rules to pre-order English for translation into subject-object-verb (SOV) languages, which is shown to be effective 
for Korean, Japanese, Hindi, Urdu, and Turkish.
\namecite{Isozaki:10b} obtain even better results in an English-to-Japanese task using only one pre-ordering rule, \ie head finalization, with a parser annotating syntactic heads.

\subsubsection{Data-driven pre-ordering} This kind of models are learned from sets of pairs $(\f,\f')$ 
where $\f$ is a source sentence and $\f'$ is its reference permutation (pre-ordering) inferred from a reference translation $\e$ via a word-level alignment.\footnote{Various heuristics have been proposed to convert a word alignment set into a sentence permutation \cite{Birch:10,Feng:10,Visweswariah:11}.} 
These approaches typically require some form of linguistic annotation of the source language, such as syntactic parse trees \cite{Xia:04,Habash:07,Li:07,Elming:09,Genzel:10,Khalilov:11j,Khalilov:11,Yang:12,Lerner:13,Jehl:14}, shallow syntax chunks \cite{Zhang:07,Crego:08} or POS labels \cite{Crego:06b,Rottmann:07,Niehues:09,Tromble:09,Visweswariah:11}.

Among the first examples of data-driven tree-based pre-ordering, 
\namecite{Xia:04} propose a method to automatically learn reordering patterns from a dependency-parsed French-English bitext, using a number of heuristics. While source-side parses are required by their method, target-side parses are optionally used to provide additional constraints during rule extraction.
\namecite{Habash:07} extracts pre-ordering rules from an Arabic-English parallel corpus dependency-parsed on the source side.
In both these works, pre-ordering rules are applied in a deterministic way to preprocess both training and test data. 
Following a discriminative modeling approach, \namecite{Li:07} train a maximum-entropy classifier to pre-order 
each node with at most 3 children in the source constituency parse, using a rich set of lexical and syntactic features.
\namecite{Lerner:13} extend this work to pre-order nodes with more children (up to 7 on either side of the head) using a cascade of classifiers: 
first, decide the order of each child relative to the head, 
then decide the order of left children and that of the right children.
As training separate classifiers for each number of children is prone to sparsity issues,
\namecite{Jehl:14} build a single logistic regression model to predict whether any two sibling nodes should be swapped or not.
Then, for each node in the tree, they search for the best permutation of all its children given
the pairwise scores produced by the model, using a depth-first procedure.
\namecite{Yang:12} treat the permutation of each node's children as a ranking problem and model it with ranking support vector machines. As an alternative to deterministic pre-ordering, 
they also propose to use the predicted source permutation to generate soft constraints for the SMT decoder: that is, a penalty that fires whenever the decoder violates the predicted pre-ordering.
A tighter integration between source pre-ordering and source-to-target translation 
is proposed by \namecite{Dyer:10}. In their approach, optimal source pre-orderings (\f') are treated as a latent variable in an end-to-end translation model
and the parameters of the tree permutation model are learnt directly from parallel data.
At test time, alternative permutations of the input tree are encoded as a \textit{source reordering forest}, which is then translated by a finite-state phrase-based translation model.

Examples of pre-ordering based on shallow syntax include \cite{Zhang:07} and \cite{Crego:08}.
In these approaches, automatically extracted chunk pre-ordering rules are used to
generate a word reordering lattice of the input sentence, which is then translated by a monotonic phrase  or n-gram based decoder.

In \cite{Costa-jussa:06}, pre-ordering is learnt by training a monolingual n-gram based SMT system at the level of word clusters.
In \cite{Tromble:09}, pre-ordering is cast as a permutation problem and solved by a model that estimates the probability of reversing the relative order of any two input words based on their distance as well as lexicalized and POS-based features.
In a related work, \namecite{Visweswariah:11} obtain smaller models and better results by learning the cost of a given input word appearing right after another, as opposed to anywhere after it (cf. source word-after-word reordering models described in Section~\ref{sec:sota-PSMT}).

\subsubsection{On the limitations of syntax-based pre-ordering}

Syntax is often regarded as the most effective way to inform reordering in translation.
However, empirical work has shown that the success of syntax-based pre-ordering methods can be severely limited by:
(i) the reachability of reference permutations when parse trees are used to constrain the pre-ordering model, and 
(ii) the quality of the parser used to learn and apply a pre-ordering model.

With regard to the constraints imposed by syntactic trees (i),
\namecite{Khalilov:12} conducted oracle pre-ordering experiments across various language pairs.
Their results consistently showed that final translation quality was highest by far when no syntactic constraint was imposed on pre-ordering (oracle string).
On the contrary, only allowing permutations of siblings of the source parse tree (oracle tree) gave the smallest improvement.
Only some of this loss could be recovered by applying specific modifications to the tree before extracting the optimal permutation (oracle modified tree).

With regard to parser accuracy (ii), \namecite{Green:09} analyzed two state-of-the-art parsers \cite{Bikel:04,Klein:03} and reported F-measures of only 55-56\% at the sub-task of detecting Arabic NP subjects in verb-initial clauses.
Similar results were observed by \namecite{Carpuat:2010} using a dependency parser \cite{Nivre:06}.
The same paper also showed that the correct pre-ordering for Arabic-English translation
could not be safely predicted even from gold standard parses,
partly due to syntactic transformations occurring during translation.
From a manual analysis of their English-German system, \namecite{Gojun:12} reported that about 10\% of the English clauses were wrongly pre-ordered, mostly due to source sentence parsing errors.
\namecite{Howlett:11} analyzed a reimplementation of the German pre-ordering method of \namecite{Collins:05}
and found that results could be affected --- or even cancelled out --- by many factors including: choice of training data, quality of the parser, as well as order of the target language model and type of reordering model used during decoding.

Rather than relying on supervised parsers trained on golden treebanks, specific parsers can be induced directly from non-annotated parallel text. 
In \cite{DeNero:11}, source sentence reorderings are first inferred from the word alignment with the target translation.
Then, a binary parsing model is trained to maximize the likelihood of source trees that can generate such reorderings.
Finally, a pre-ordering model is trained to permute each node in the tree.
Evaluated on the English-Japanese language pair, this method almost equals the performance of a pre-ordering method based on a supervised parser.
\namecite{Neubig:12} follow a similar approach but build a single ITG-style pre-ordering model treating the parse tree as a latent variable.
In the target self-training method of \namecite{Katz-Brown:11}, 
a baseline treebank-trained parser is used to produce n-best parses of a parallel corpus' source side.
Then, the parses resulting in the most accurate pre-ordering after application of a dependency-based pre-ordering rule set \cite{Xu:09} are added to the treebank data and used to re-train the baseline parser.

\subsubsection{Post-ordering}
A somewhat smaller line of research has instead treated reordering as post-processing.
In \cite{Bangalore:00b,Sudoh:11}, target words are reordered after a monotonic translation process.
Other work has focused on rescoring a set of n-best translation candidates produced by a regular PSMT decoder
--- for instance by means of POS-based reordering templates \cite{Chen:06a} or word-class specific distortion models \cite{Gupta:07}.
\namecite{Chang:07} use a dependency tree reordering model to generate \textit{n} alternative orders for each 1-best sentence produced by the SMT system. Each set of \textit{n} sentence reorderings is then reranked using a discriminative model trained on word bigram features and standard word reordering features (\ie distance or orientation between consecutively translated input words).

Focusing on Japanese-to-English translation, \namecite{Sudoh:11,Sudoh:13j} proposed to `translate' foreign-order English into correct-order English using a monolingual phrase-based \cite{Sudoh:11} or syntax-based \cite{Sudoh:13j} SMT system trained for this specific subtask.%
\footnote{Note the similarity to the pre-ordering approach of \namecite{Costa-jussa:06}, except that here the monolingual SMT process is applied to the target language after a monotonic translation phase.}
The underlying motivation is that, while English-to-Japanese is well handled by pre-ordering with the aforementioned head-finalization rule \cite{Isozaki:10b}, it is much harder to predict the English-like order of Japanese constituents for Japanese-to-English translation.
Post-ordering addresses this issue by generating head-final English (HFE) sentences that are used to create a HFE-to-English parallel corpus.
\namecite{Goto:12,Goto:13j} solve post-ordering by parsing the HFE sentences into binary trees annotated with both syntactic labels and ITG-style monotone/swap labels.
\namecite{Hayashi:13} improve upon this work with a shift-reduce parser that efficiently integrates non-local features like n-grams of the post-ordered string.

Also related to post-ordering is the work on right-to-left or \textbf{reverse decoding} by \namecite{Watanabe:02,Finch:09}, and \namecite{Freitag:13}.
Here, the target sentence is built up from the last word to the first, thereby altering language model context and reordering search space.
\namecite{Finch:09} obtain best results on a wide range of language pairs by combining the outputs of standard and reverse decoding systems.

\section{Evaluating Word Reordering in Statistical Machine Translation}
\label{sect:evaluation}

Since there are innumerable ways to correctly render a source sentence's meaning in the target language, automatically evaluating translation quality is a complex problem.
Generally, SMT systems are  judged by the extent to which their outputs resemble a set of reference translations produced by different human translators.
Despite relying on a very rough approximation of language variability, this approach provides SMT researchers with fast automatic metrics that can guide, at least in part, their steps towards improvement.
Besides, fast evaluation metrics are used to automatically tune SMT feature weights on a development corpus, for instance by means of minimum error rate training procedures \cite{Och:03c}. 
The design of MT evaluation metrics correlating with human judgements is an active research area.
Here we briefly survey two widely used general-purpose metrics, BLEU and METEOR, and then describe in more detail a number of reordering-specific metrics.

\subsection{General-purpose metrics}
\textbf{BLEU} 
\cite{Papineni:01} is a lexical match based score that represents the de facto standard for SMT evaluation.
Here, proximity between candidate and reference translations is measured in terms of overlapping word $n$-grams, with $n$ typically ranging from 1 to 4.
For each order $n$ a {\em modified precision} score (see \namecite{Papineni:01} for details) is computed on the whole test set and combined in a geometric mean. The resulting score is then multiplied by a {\em brevity penalty} that accounts for length mismatches between reference and candidate translations.
\namecite{Al-Onaizan:06} use BLEU to measure word order similarity between two languages:
that is, by computing the BLEU score between the original target sentence $\e$ and a source-like permutation of $\e$.
Using $n$-grams, though, is a limited solution to the problem of word ordering evaluation: 
First, because only exact surface matches are counted, without any consideration of morphology or synonymy. 
Second, because the absolute positioning of words in the sentence is not captured, but only their proximity within a small context.

The former issue is addressed to some extent by \textbf{METEOR} 
\cite{Banerjee:05}, which relies on language-specific stemmers and synonymy modules to go beyond the surface-level similarity.
%
As for word order, METEOR treats it separately with a {\em fragmentation penalty} proportional to the smallest number of chunks that the hypothesis must be divided into to align with the reference translation. 
This quantity can be interpreted as the number of times that a human reader would have to `jump' 
between words to recover the correct translation order.
However, no distinction is made between short and long-range reordering errors.

The weakness of BLEU and METEOR with respect to word order was demonstrated by \namecite{Birch:10}
with a significant example that we report in Table~\ref{tab:Birch}.
For simplicity, the example assumes that the reference order is monotonic and that
hypotheses and reference translations contain exactly the same words.
According to both metrics, hypothesis (a) is worse than (b), although 
in (a) only two adjacent words are swapped while in (b) the two halves of the sentence are swapped.

\begin{table}[t]\centering
\hspace{-2mm}
\begin{minipage}{0.62\linewidth}
\includegraphics[width=\textwidth]{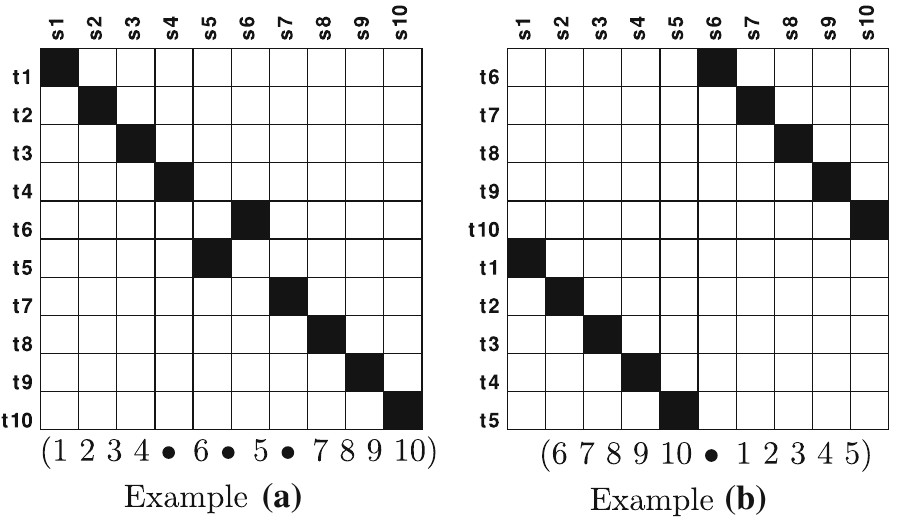}
\end{minipage}
\hspace{1mm}
\begin{minipage}{0.25\linewidth}
\begin{small}\begin{tabular}{@{\ }l@{\ \ } c @{\ \ } c @{\ }}
\hline
Example:	& \bf(a) & \bf(b) \\
\hline
\textsc{bleu}	  & 61.80 & 81.33 \\
\textsc{meteor} & 86.91 & 92.63 \\
\hline
\end{tabular}\end{small}
\end{minipage}
\caption{\label{tab:Birch} Two example alignments and their respective BLEU and METEOR scores, assuming that the reference alignment is  monotonic.
The permutation resulting from the hypothesis alignment is reported under each matrix, where bullet points represent jumps between non-sequential indices.
Taken from \protect\namecite{Birch:11:thesis}.}
\end{table}

\subsection{Reordering-specific metrics}

To overcome the aformentioned limitations, \namecite{Birch:10} propose to directly measure the similarity between the reorderings needed to reach the
reference translations from the source sentence and those applied by the decoder to produce the candidate translation. 
In practice, this is done by first converting word alignments to permutations using simple heuristics to handle null and multiple alignments,
and then computing a permutation distance among the resulting permutations.
Among various metrics proposed in the paper, the square root of the Kendall's Tau was shown to be reliable and highly correlated with human judgements.

The normalized Kendall's Tau distance $K$ is originally a measure of disagreement between rankings.
Given a set of $n$ elements and two permutations $\pi$ and $\sigma$, the $K$ distance corresponds to the number of discordant pairs (i.e. pairs of elements whose relative order differs in the two permutations) normalized by the total number of ordered element pairs:
 \[ K(\pi,\sigma) = \frac{\sum_{i=1}^{n}\sum_{j=1}^{n} \mathbf{d}(i,j)}{\frac{1}{2}n(n-1)}	 
 \quad \textrm{where} \quad
 \mathbf{d}(i,j) = \left\{ \begin{array}{ll}
          1 & \mbox{if $\pi_{i}<\pi_{j}$ and $\sigma_{i}>\sigma_{j}$}\\
          0 & \mbox{otherwise}\end{array} \right. \] 
\noindent
\namecite{Birch:10} further suggest to extract the square root of $K$ to obtain a function that is more discriminative on lower distance ranges, 
i.e. for translations that are closer to the reference word ordering.
Finally, the \textbf{Kendall Reordering Score (KRS)} --- a positive measure of quality ranging from 0 to 1 --- 
is computed by subtracting the latter quantity from one, and by multiplying the result by a brevity penalty ($BP$) 
that accounts for length mismatches between reference and candidate translations:
\[  \textsl{KRS}(\pi,\sigma) = (1-\sqrt{K(\pi,\sigma)}) \cdot BP \]
\noindent
The $BP$ definition corresponds to that of BLEU  \cite{Papineni:01}  with the difference that, for KRS, it is computed at the sentence level.
In case of multiple references, the one that yields the highest score for each test sentence is retained.
Finally, the average of all sentence-level KRS scores gives the global KRS of the test set. 
The linear interpolation of KRS and BLEU (\textbf{LRscore}) can be successfully used to
optimize the feature weights of a PSMT system, leading to translation outputs that are preferred by human annotators according to \namecite{Birch:11}.
%

In a related work, \namecite{Bisazza:13} observe that some word classes, like verbs, are typically more important than others to determine the general structure of a sentence.
Hence, they develop a word-weighted KRS variant that is more sensitive to the positioning of specific input words.
Assuming that each input word $f_i$ is assigned a weight $\lambda_i$,  the original KRS formula is modified as follows:
 \[
 \mathbf{d_\lambda}(i,j) = \left\{ \begin{array}{ll}
       \lambda_i {+} \lambda_j & \mbox{if $\pi_{i}<\pi_{j}$ and $\sigma_{i}>\sigma_{j}$}\\
          0 & \mbox{otherwise}\end{array} \right. \] 
For their evaluation of long reordering errors in Arabic-English and German-English,
\namecite{Bisazza:13} set the weights to 1 for verbs and 0 for all other words to only capture verb reordering errors. 
The resulting metric, \textbf{KRS-V}, rates a translation hypothesis as perfect when the translations of all source verbs are located in their correct position, regardless of the ordering of other words.

In a different approach called \textbf{RIBES}, \namecite{Isozaki:10} propose to directly measure the reordering occurring between the words of the hypothesis and those of the reference translation, thereby eliminating the need to word-align input and output sentence. 
A limitation of this approach is that only identical words contribute to the score.
As a solution, the permutation distance is multiplied by a word precision score
that penalizes hypotheses containing few reference words.
Nevertheless, the resulting metric assigns different scores to hypotheses that differ in their lexical choice, but not in their word reordering.

\namecite{Talbot:11} introduce yet another reordering-specific metric, called fuzzy reordering score (\textbf{FRS}) which, like the KRS,  is independent from lexical choice and measures the similarity between a sentence's reference reordering and the reordering produced by an SMT system (or by a pre-ordering technique).
However, while \namecite{Birch:10} employed the Kendall's Tau between the two sentence permutations,
\namecite{Talbot:11} count the smallest number of chunks that the hypothesis permutation must be divided into to align with the reference permutation. This corresponds precisely to the fragmentation penalty of METEOR except that the alignment is performed between permutations and not between translations. 
Like METEOR, FRS makes no difference between short and long-range reordering errors (cf. Table~\ref{tab:Birch}).

\COMMENT{A lightweight evaluation framework for machine translation reordering D Talbot:
- compared to Birch: slightly different heuristics for alignment->permutation:
If multiple source tokens are aligned to a single target word or span we ignore the ordering within these source spans; this is indicated by braces in Table 2. We place unaligned source words immediately before the next aligned source word or at the end of the sentence if there is none
- employ a dedicated gold standard composed of very conservative (i.e. literal) translations and manual word alignments generated by the same translators (ie  created with a bias towards simple phrasal reordering)
- They show that fragmentation penalty is better correlated with human judgments of translation quality than KendallÕs (but not its square root!).
}
 
\namecite{Stanojevic:14:EMNLP} argue for a hierarchical treatment of reordering evaluation, where word sequences can be grouped recursively into larger blocks.
To this end, they factorize the output-reference reordering into a Permutation Tree \cite{Zhang:07:PET}, 
whose nodes represent atomic permutations. 
Given this factorization, the counts of monotone \mbox{(1 2)} versus other permutation nodes --- \mbox{(2 1)}, \mbox{(3 1 4 2)}, etc. ---
are used as features in a linear model of translation quality (\textbf{BEER}) trained to correlate with the human ranking of a set of MT system outputs.
With reference to Table~\ref{tab:Birch}, the permutation trees of both hypotheses (a) and (b) would contain only one swapping node leading to the same reordering score.
\namecite{Stanojevic:14:SSST} extend this work with a stand-alone reordering metric that considers all possible tree factorizations of a permutation (permutation forest) and that gives recursively less importance to lower nodes in the tree (i.e. covering smaller spans).
Hierarchical permutation metrics are shown to better correlate with human judgements than string-based permutation metrics like the Kendall's Tau distance $K$.

\section{Reordering Phenomena in Natural Languages}
\label{sect:phenomena}

Understanding the complexity of reordering in a given language pair is key to selecting the right SMT models and to improving them.
To date, word reordering phenomena in natural languages have mainly been analyzed from a quantitative perspective \cite{Birch:08,Birch:09}.
While measuring the \textit{amount} of reordering is certainly important, understanding which \textit{kinds} of reordering occur in a given language pair is also essential.
To this end, we present a qualitative analysis of word reordering based on linguistic knowledge.
More specifically, we draw on a large body of syntactic information collected by linguists from more than 1500 languages,
and systematized in the World Atlas of Language Structures (WALS) \cite{WALS-2011}\footnote{\url{http://wals.info}}.

Following the seminal work of language typologist Matthew S. Dryer, we describe the word order profile of a language
by the canonical orders of its constituent sets (word order features).
The resulting language pair classification is primarily based on the order of subject, object and verb,
and further refined according to the order of several other element pairs, such as noun-adjective, verb-negation, etc.
We then compare the word order features of several languages that were studied in the SMT field, 
and show that empirical results generally confirm the existing theoretical knowledge.

\subsection{A qualitative analysis}
\label{sec:order-analysis}

The amount of word reordering found in a language pair is known to be a good predictor of SMT performance.
\namecite{Birch:08} considered three variables --- reordering quantity, morphological complexity and historical relatedness --- and
found the first to have the highest correlation with the BLEU scores of a standard PSMT system 
on a sample of 110 European language pairs.
\namecite{Birch:09} further analyzed the distribution of different reordering widths in Arabic-English and Chinese-English,
and the ability of two SMT approaches to model them.
They found that
the PSMT approach is more suitable for language pairs where most reordering is local (Arabic-English), while
the hierarchical approach is stronger when medium-range reorderings are dominant (Chinese-English).
Still, both PSMT and HSMT failed to capture most of the long-range reorderings found in the reference corpora.

These findings are indeed relevant to our work, but we believe there is also much to learn from theoretical linguistic knowledge. 
Moreover, a quantitative analysis can suffer from noise in the data, typically originating from automatic word alignments.
\namecite{Birch:09} used manual word alignment in their study, but this kind of resource is available only for very few language pairs.
Noise can also be due to what we can call \textit{optional} reordering: human translators often choose to restructure the sentence according to genre conventions or to their personal style, even when this is not required by the target language grammar.
Here is an example:
\renewcommand{\arraystretch}{1.3}
\begin{center}
\begin{small} \sf
\begin{tabular}{p{.8\textwidth}}
\specialrule{.8pt}{1pt}{1pt}
\multicolumn{1}{c}{\it Arabic sentence:}\\
\multicolumn{1}{r}{\small \< w .tm'an bw^s ($55$ snT) Al.s.hAfyyn qbyl m.gAdrth Albyt AlAby.d AlY Anh y^s`r bAnh fy > } \\
\multicolumn{1}{r}{\small \< .hAl $\mrq\mrq$rA'i`T$\mlq\mlq$ w .s.hT $\mrq\mrq$jydT jdA$\mlq\mlq$. > } \\
\specialrule{.8pt}{1pt}{1pt}
\multicolumn{1}{c}{\it Literal translation:}\\
Bush, aged 55, assured journalists \underline{before leaving the White House} that he felt ``great'' and that his health was ``very good''.\\
\specialrule{.8pt}{1pt}{1pt}
\multicolumn{1}{c}{\it Human translation:}\\
\underline{Before leaving the White House}, Bush, aged 55, assured journalists that he felt ``great'' and that his health was ``very good''.\\
\specialrule{.8pt}{1pt}{1pt}
\end{tabular}
\end{small}
\end{center}
\renewcommand{\arraystretch}{1}
As also noted by \namecite{Fox:02}, this kind of reordering is not strictly necessary to produce accurate and fluent translations, but its occurrence in parallel corpora affects the automatic reordering measures.

On the contrary, a qualitative analysis can profit from the extensive work done by linguists and grammaticians to abstract the fundamental properties of a language.
In this section we draw largely on \namecite{Dryer:07} and on the sections of WALS devoted to word order 
(\namecite{WALS-2011-81}, ch. 81-97, 143-144).


\subsection{Word order profiles}
\label{sec:order-profiles}

The word order profile of a language is determined by the canonical order of its constituent sets, or word order features.
In general, the basic or canonical order of a constituent set can be established by criteria of 
frequency (the most common), distribution (the one with the least restricted usage) or pragmatics (the neutral one) \cite{Dryer:07}.
Although some languages are said to have free (or flexible) order, it is often possible to detect one that is dominant and neutral.
Consider for instance English, a subject-verb-object (SVO) language where other orders are used, but only to achieve specific emphasis or topicalization effects:
\eenumsentence{
\setlength{\itemsep}{1mm}\setlength{\parskip}{1mm}%
\item {\sf I saw the cat.}
\item {\sf The cat, I saw.}
}
However, there exist cases where no particular order can be defined as dominant.
An example of mix-ordered constituent set in English is the pair noun and genitive:
\eenumsentence{
\setlength{\itemsep}{1mm}\setlength{\parskip}{1mm}%
\item {\sf the tail of the cat}
\item {\sf the cat's tail}
}

Based on \namecite{Dryer:07} and on the availability of data points in the WALS,
we have established a set of 13 core features to determine the word order profile of a language.
For the purpose of describing word order differences between language pairs, we have divided the features into two broad categories:
clause-level and phrase-level%
\footnote{In this section, phrase is used in its traditional syntactic sense --- \ie a group of words forming a constituent --- 
as opposed to the notion of data-driven phrase adopted by phrase-based SMT.}.
An English example for each feature is provided in Table~\ref{tab:order-features}.

\subsubsection{Clause-level order features}

\begin{itemize}
\setlength{\itemsep}{1mm}\setlength{\parskip}{1mm}%
\justifying
\item \textbf{Subject, Object, Verb} [WALS feature 81A]  \hfill \\
The first and most important feature is the ``ordering of subject, object, and verb in a transitive clause, more specifically declarative clauses in which both the subject and object involve a noun (and not just a pronoun)'' \cite{WALS-2011-81}.
For instance, English and French are SVO languages, while Turkish is SOV.
The distribution of main word order types in a large sample of world languages is given in Table~\ref{tab:worder-distrib}.
This feature is often used alone to denote the word order profile of a language, 
because it can be a good predictor of several other features.

\begin{table}[h]
\centering \small
\begin{tabular}{| c | r @{} r @{\ \ \ }|}
\hline
\bf Order & \multicolumn{2}{ c |}{\bf Languages} \\
\hline
	SOV &  565 & 41\% \\
	SVO &	 488	& 35\% \\
	VSO &	 95 & 7\%  \\
	VOS &	 25 &	 2\% \\
	OVS &	 11 &	 1\% \\
	OSV &	 4  &	 \textless 1\%  \\  
	mixed/no-dominant & 189 & 14\% \\
\hline
	total sample size	& 1377 & \\
\hline
\end{tabular}
\caption{\label{tab:worder-distrib}
The distribution of main word order types (Subject, Object, Verb) in the world languages. 
From the World Atlas of Language Structures, chapter 81 (Dryer 2011). }
\vspace{-2mm}
\end{table}

\item \textbf{Oblique or Adpositional Phrase} [84A] \hfill \\
This feature refers to the position of a phrase functioning as adverbial modifier of the verb, relative to the position of object and verb. For instance, English is VOX because it places oblique phrases after verb and object.

\item \textbf{Noun and Relative Clause} [90A]  \hfill \\
Order of the relative clause with respect to the noun it modifies.

\item \textbf{Adverbial Subordinator and Subordinate Clause} [94A]  \hfill \\
Subordinators are used to link adverbial subordinate clauses to the main clause.
They can take the form of verbal suffixes or separate words, such as the English subordinating conjunctions `when' and `because'.

\item \textbf{Polar Question Particle} [92A]  \hfill \\
In many languages, polar (yes/no) questions are signaled by specific particles. 
This feature denotes their position in the sentence (not defined for English).

\item \textbf{Content Question Phrase} [93A]  \hfill \\
Content questions are characterized by the presence of an interrogative word or phrase (e.g. `who', `which one').
In some languages, like English, these are always placed at the beginning of the sentence.
In some others, like Turkish, they take the position of the constituent they replace:
for instance, the word \textit{`ne}/what' replacing the object naturally occurs between subject and verb.

\item \textbf{Negation and Verb} [143A]  \hfill \\
Order of the negative word or morpheme\footnote{Unlike the WALS, we do not distinguish between negative words and affixes for this feature.} 
with respect to the \textit{main} verb. 
Note that more than one word or morpheme may be necessary to express negation (\eg \textit{`ne ... pas'} in French).

\end{itemize}

\subsubsection{Phrase-level order features}

\begin{itemize}
\setlength{\itemsep}{1mm}\setlength{\parskip}{1mm}%

\item \textbf{Noun and Adpositions} [WALS feature 85A] \hfill \\ 
Whether a language uses mainly prepositions or postpositions.

\item \textbf{Noun and Genitive} [86A] \hfill \\
Order of genitive or possessor noun phrase with respect to the head noun.
 
\item \textbf{Noun and Adjective} [87A] \hfill \nopagebreak[4] \\ 
Order of adjectives with respect to the noun they modify.

\item \textbf{Noun and Demonstrative} [88A] \hfill \\
Order of demonstrative words (e.g. \textit{this}, \textit{that}) or affixes with respect to the noun they modify.

\item \textbf{Noun and Numeral} [89A] \hfill \\
Order of cardinal number words with respect to the noun they modify.

\item \textbf{Adjective and Degree Word} [91A] \hfill \\
Order of degree words (e.g. \textit{very}, \textit{more}) with respect to the adjective they modify.

\end{itemize}

\customspacestyle{-12}

\renewcommand{\arraystretch}{1.3}
\begin{table}[h]
\centering \small
\begin{tabular}{|@{\ }c@{\ } | l@{\ } | l@{\ \ } l@{\ \ } l@{\ \ } l@{\ \ } l@{\ \ } l@{\ \ } l@{\ }|}
\cline{3-9} 
\multicolumn{2}{@{}c@{\ }|}{} &		\multicolumn{3}{c}{\it \hspace{-2mm} Indo-European} & \hspace{-5mm} \it Afro-Asiatic & \it \  Altaic & \it \hspace{-0.5mm}  Japanese & \it \hspace{-1.5mm} Sino-Tibet. \\
\multicolumn{2}{@{}c@{\ }|}{} &		\multicolumn{2}{c}{\it \ Germanic} & \multicolumn{1}{c}{\it \hspace{-9mm} Romance} &\hspace{-1mm} \it Semitic & \it \  Turkic & \it \hspace{-0.5mm} Japanese & \it \hspace{-0.5mm} Chinese \\
\cline{1-2}
\multicolumn{2}{|c|}{\bf Features}	&	\bf English	&	\bf German	& \bf French &	\bf Arabic	&	\bf Turkish	& \bf Japanese &	\bf Chinese	\\
\hline\hline  

\multirow{13}{*}{\begin{sideways} \bf Clause-level \end{sideways}}  
	& Subject,Object,Verb &	S-V-O	&	S-V-O/	& S-V-O &	V-S-O/	&	S-O-V	& S-O-V & S-V-O	\\
	& {\scriptsize \sf [Tom] [chases] [Jerry]} & & S-O-V & & S-V-O* & & & \\
\cline{2-9} 
	& Oblique Phrase	&	V-O-X	&	mixed		& V-O-X &	V-O-X				&	X-O-V	& X-O-V & X-V-O	\\
	& {\scriptsize \sf [chases] [Jerry] [with a stick]} & & & & & & & \\
\cline{2-9} 
	& Noun,RelClause			& N-Rel	&	N-Rel	&	N-Rel	&	N-Rel*	&	Rel-N	& Rel-N	& Rel-N \\ 
	& {\scriptsize \sf [a stick] [that he stole]} & & & & & & & \\
\cline{2-9} 
	& Subordinator,Clause	&	Sub-C	&	Sub-C	& Sub-C	&	Sub-C		&	C-Sub/	&	C-Sub & mixed**	\\
	& {\scriptsize \sf [because] [he was hungry]} & & & & & Sub-C & & \\
\cline{2-9} 
	& PolarQuest.Particle		& \it none &	\it none		& initial	&	initial	 & final	& final &	final	\\
	& {\scriptsize $\varnothing$ \sf [did Tom steal it?]} & & & & & & & \\
\cline{2-9} 
	& ContentQuest.Phrase		& initial	&	initial &	initial	&	initial*	&	other & other & other \\
	& {\scriptsize \sf [what] [did Tom steal?]} & & & & & & & \\	
\cline{2-9} 
	& Negation,Verb		&  Neg-V	&	Neg-V/ &	\customspacestyle{-90}Neg-V-Neg/	\customspacestyle{-15} &	Neg-V	&	V-Neg & V-Neg & Neg-V \\
	& {\scriptsize \sf he did [not] [steal]} & & V-Neg & V-Neg & & & & \\	
\hline\hline  

\multirow{11}{*}{\begin{sideways} \bf Phrase-level \end{sideways}}
	& Noun,Adpositions		& Adp-N	&	Adp-N			&	Adp-N	& Adp-N				& N-Adp	&	N-Adp	& N-Adp/ \\
	& {\scriptsize \sf [with] [a stick]} & & & & & & & Adp-N \\	
\cline{2-9} 
	&	Noun,Genitive				& N-Gen/	&	N-Gen			& N-Gen	& N-Gen				& Gen-N	& Gen-N & Gen-N \\
	& {\scriptsize \sf [Tom's] [stick]} & Gen-N & & & & & & \\	
\cline{2-9} 
	&	Noun,Adjective			& A-N	&	A-N			& N-A	& N-A				& A-N	& A-N & A-N \\
	& {\scriptsize \sf [hungry] [Tom]} & & & & & & & \\	
\cline{2-9} 
	&	Noun,Demonstrative	& Dem-N &	Dem-N	& Dem-N	& Dem-N		& Dem-N	& Dem-N & Dem-N \\
	& {\scriptsize \sf [this] [stick]} & & & & & & & \\	
\cline{2-9} 
	&	Noun,Numeral			& Num-N	&	Num-N		& Num-N	& Num-N		& Num-N	& Num-N & Num-N \\
	& {\scriptsize \sf [two] [sticks]} & & & & & & & \\	
\cline{2-9} 
	&	Adjective,DegreeW.& Deg-A &	Deg-A				& Deg-A	& A-Deg				& Deg-A	& Deg-A & Deg-A \\
	& {\scriptsize \sf [very] [hungry]} & & & & & & & \\	
\hline\hline
\multicolumn{2}{|c|}{\bf Feature}	&	\bf English	&	\bf German	& \bf French &	\bf Arabic	&	\bf Turkish	& \bf Japanese &	\bf Chinese	\\
\hline
\end{tabular}
\renewcommand{\arraystretch}{1.1}
\caption{\label{tab:order-features}
The word order profile of seven world languages. Language family and genus (Dryer 1989) are indicated in the header's first and second row, respectively. \
Sources: the World Atlas of Language Structures (Dryer and Haspelmath 2011), 
(*) authors' knowledge, and (**) (Li 2008).
} 
\end{table}

\normalspacestyle

\nocite{Dryer:89}

\subsubsection{Language sample}

For our study, we have chosen seven widely spoken languages.
These are English, German, French, Arabic (Modern Standard), Turkish, Japanese and Chinese (Mandarin).
Mainly based on the WALS, we have summarized the word order feature values for all these languages in Table~\ref{tab:order-features}.
Whenever possible, features were assigned one (or two) values corresponding to the dominant order(s) in that language. When no particular order was given as dominant we marked it as `mixed'.

The main word order of German and Arabic deserves a special mention.
In German, the positioning of subject, object and verb is syntactically determined: 
main clauses without auxiliary verb are SVO, while
subordinate clauses and clauses containing an auxiliary are SOV.
A further complication, not marked in Table~\ref{tab:order-features}, is that the German finite verb must be placed in second position, in which case the pattern becomes S\textit{Aux}OV, with the object intervening between auxiliary and main verb.
As regards Arabic, while the WALS classifies Modern Standard Arabic as VSO, the corpora typically used in SMT show a very mixed distribution of VSO and SVO clauses.%
\footnote{VOS order is also admitted in Arabic, but only in specific contexts (\eg when the object is expressed by a pronoun).}
\namecite{Carpuat:12} examined the Arabic-English Treebank 
and found that, when the subject is expressed, it follows the verb in 70\% of the cases,
but precedes it in 30\%.
Similarly, in the Pennsylvania Arabic Treebank, 
they found an order distribution of 67\% VS and 33\% SV.
%
%
Besides frequency, it can be noted that the SVO sentences attested in these corpora are in general pragmatically neutral.
We conjecture that this variability in Modern Standard Arabic may be due to the effect of spoken language varieties such as Egyptian, Gulf, Kuwaiti, Iraqi (all listed as SVO by the WALS), and Syrian (listed as VSO/SVO).
For these reasons, we classify Arabic as a mixed VSO/SVO language.

It is worth noting that our seven-language sample covers the main word order types of the large majority of the world languages: namely SOV, SVO and VSO (see Table~\ref{tab:worder-distrib}).


\subsection{Word order differences}
\label{sec:order-differences}

Linguistically motivated word order profiles can be very helpful to anticipate the kind of word reordering problems that an SMT system will have to face.
Clearly, these will also vary in relation to the text genre (written news, speeches, etc.) and to the translation's style and degree of literality. 
However, we can reasonably expect the syntactic properties of two languages to determine the general reordering characteristics of that pair.

We will now analyze the reordering characteristics of seven language pairs:
English paired with the other six languages presented in Table~\ref{tab:order-features}, as well as the French and Arabic pair.
%
To this end, we propose the following analysis procedure.
As a first indication of reordering complexity, we look at
the main word order feature (subject, object, verb).
A difference at this level typically results in poor SMT performances.
Then, we count the total number of discordant features. 
To simplify, if a particular element does not exist in a language (\eg polar question particles in English) we  count \textit{zero} difference for that feature, and when one of the languages has a mixed order we count a \textit{half} difference.
We insist, however, on the qualitative nature of our analysis: numbers are only meaningful in combination with the list of specific discordant features, as these have different impact on word reordering.
In particular, we find it essential for SMT to distinguish between clause-level and phrase-level differences (\textbf{CDiff} and \textbf{PDiff}) 
because the former account for most longer-range word movements, and the latter for the shorter.
Thus, a language pair with only phrase-level discordant features
is likely to be suitable for a PSMT approach, where reordering is managed through local distortion or inside translation units.
On the contrary, the presence of many clause-level differences typically calls for a tree-based solution, either at preprocessing or at decoding time.
As we will see, some pairs lay on the borderline, with only one or few clause-level differences.
Finally, it should be noted that, even among features of the same group, some have more impact on SMT than others due to their frequency or to the average length of their constituents. For instance, the order of noun and genitive is more important than that of adjective and degree word.

\begin{description}
\item[English and German]  [ Main order: different;\ \ CDiff=1.5;\ \ PDiff=0.5 ] \hfill \\ 
The main word order of German is SVO or SOV according to the syntactic context (cf. Section~\ref{sec:order-profiles}).
German also differs from English with respect to the position of oblique phrases and that of the negation: both fixed in English but mixed in German.
At the phrase level, German predominantly places the genitive after the noun, while English displays both orders.

Thus, despite belonging to the same family branch (Indo-European/Germanic), this pair displays complex reordering patterns.
Indeed, German-English reordering has been widely studied in SMT and is still an open topic.
%
At the Workshop on Statistical Machine Translation 2014 \cite{WMT:14}, a syntax-based string-to-tree SMT approach \cite{Williams:14} was winning in both language directions (official results excluding online systems).
At the International Workshop on Spoken Language Translation 2014 \cite{IWSLT:14},
the best submission was a combination of PSMT with POS- and syntax-based preordering \cite{Slawik:14}, string-to-tree syntax-based SMT and factored PSMT \cite{Birch:14}.





\item[English and French] [ Main order: same;\ \ CDiff: 0.5;\ \ PDiff: 1.5 ] \\
Most clause-level features have the same values in French as in English, 
except for the negation which is typically expressed by two words in French: one preceding and one following the verb.\footnote{Pre-verbal negation can be omitted in colloquial French.}
At the phrase level, differences are found in the location of genitives and adjectives.
Thus, English and French have very similar clause-level orders, but reordering is abundant at the local level.

This is a case where reordering is mostly well handled by string-based PSMT.  
As a reference, the three top English-to-French WMT14 systems (official results excluding online systems), were all phrase-based. A similar trend was observed in the French-to-English track.
%

\item[English and Arabic] [ Main order: different;\ \ CDiff: 0.5;\ \ PDiff: 2.5 ] \\
The dominant Arabic order is VSO, followed by SVO (cf. Section~\ref{sec:order-profiles}).
Apart from this important difference, all other clause-level features agree between Arabic and English.
At the phrase level, differences are found in genitives, adjectives and degree words.

As a result, reordering is overwhelmingly local but few crucial long-range reorderings also regularly occur.
Thus, this pair is challenging for PSMT but, at the same time, not well suited for a tree-based approach.
As shown by \namecite{Zollmann:08} and \namecite{Birch:09}, PSMT performs similarly or better than HSMT for the Arabic-to-English language pair.
However, HSMT was shown to better cope with the reordering of VSO sentences \cite{Bisazza:phdthesis:13}.
Pre-ordering of Arabic VSO sentences for translation into English has proved to be a particularly difficult task \cite{Green:09,Carpuat:2010} and has inspired work on hybrid pre-ordering where multiple verb pre-orderings are fed to a PSMT decoder \cite{Bisazza:2010:WMT,Andreas:11}, see also Section~\ref{sec:sota-preproc}.




\item[English and Turkish] [ Main order: different;\ \ CDiff: 5.5;\ \ PDiff: 1.5 ] \\
Turkish is a good example of head-final language, except for the fact that it can employ both clause-final and clause-initial subordinators.%
\footnote{In Turkish, non-finite subordinate clauses are typically placed before the main clause and linked to it by a clause-final subordinator (\eg \textit{ra\v{g}men/although}), whereas finite subordinate clauses can be placed after the main clause and introduced by a clause-initial subordinator (\eg \textit{ama/but}).
The former is dominant in written language.}
As a result, almost all clause-level features are discordant in this pair.
At the phrase level, Turkish mainly differs from English for the use of postpositions instead of prepositions.

Among our language pairs, this is one of the most difficult to reorder for an SMT system. 
The complex nature of its reordering phenomena suggests a good fit for tree-based SMT approaches, and
indeed HSMT was shown to significantly outperform PSMT between Turkish and English in both language directions \cite{IWSLT:12,Yilmaz:13}.
However, state-of-the-art SMT quality in this language pair is still very low, 
mostly due to the agglutinative nature of Turkish which makes it difficult to tear apart word reordering issues from rich morphology issues. 
Attempting to address both issues in an English-to-Turkish factored PSMT system, 
\namecite{Yeniterzi:10} pre-process the parsed English side with a number of syntax-to-morphology mapping rules and costituent pre-ordering rules dealing with local and global reordering phenomena respectively.
Only the former, though, resulted in better translation quality.


\item[English and Japanese] [ Main order: different;\ \ CDiff: 6;\ \ PDiff: 1.5 ] \\
Japanese is the prototypical example of head-final language. 
In this pair all clause-level features are discordant, while at the phrase level, Japanese differs from English for the use of postpositions and the strictly head-final genitive construction.

This pair, like the previous one, is extremely challenging for PSMT due to the hierarchical nature of its reordering phenomena and the high frequency of long-range word movements.
Indeed, translation between English and Japanese has spurred a remarkable amount of work on pre-ordering, post-ordering and decoding-time reordering. 
In 2013 the PatentMT evaluation campaign of the NTCIR conference \cite{Goto:13:patentMT} saw rule-based and hybrid systems largely outperform the purely statistical ones in Japanese-to-English. 
The highest-ranked SMT submission was actually a combination of three SMT systems including: 
a baseline PSMT method, 
a rule-based pre-ordering method, 
and a post-ordering method based on string-to-tree syntax-based SMT \cite{Sudoh:13}. 
Interestingly, the trends were different in the opposite translation direction, English-to-Japanese, where all rule-based MT systems were significantly outperformed by a PSMT system that performed pre-ordering of the English input with few manual rules for head finalization based on dependency parse trees \cite{Sudoh:13}.


\item[English and Chinese] [ Main order: same;\ \ CDiff: 3.5;\ \ PDiff: 1 ] \\
Despite belonging to the same main order type,
these two languages differ in the positioning of oblique phrases, relative clauses, interrogative phrases and subordinating words.\footnote{Subordinating words in Chinese can occur at the beginning of the subordinate clause, at its end, or even inside it \cite{Li:08}.}
Moreover, word order variations are quite common in Chinese to mark the \textit{topic} of a sentence, i.e. what is being talked about.
Comparing the two languages at the phrase level, we find partial disagreement in the use of genitive and adpositions (Chinese has both prepositions and postpositions).

Thus, this pair too is characterized by very complex reordering, hardly manageable by a PSMT system.
This is confirmed by a number of empirical results showing that tree-based approaches (particularly HSMT) consistently outperform PSMT in Chinese-to-English evaluations \cite{Zollmann:08,Birch:09}.
It is worth noting that translation between Chinese and English has been the main motivation and test bed for the development of HSMT. 



\item[French and Arabic] [ Main order: different;\ \ CDiff: 1.5;\ \ PDiff: 1 ] \\
At the clause level, this pair differs in main word order (SVO versus VSO or SVO) like the English-Arabic pair, but also in the order of negation and verb.
On the other hand, phrase-level order is notably more similar, with only one discordant feature of minor importance (adjective and degree word).

Less research was published on this language pair. 
Nevertheless, \namecite{Hasan:08} and \namecite{Schwenk:2009} chose a PSMT approach to experiment with an Arabic-to-French task.

\end{description}

\begin{figure}[p]
\begin{center}

English and German:
\vspace{4mm}

\hspace{-11mm}
\includegraphics[width=.48\textwidth]{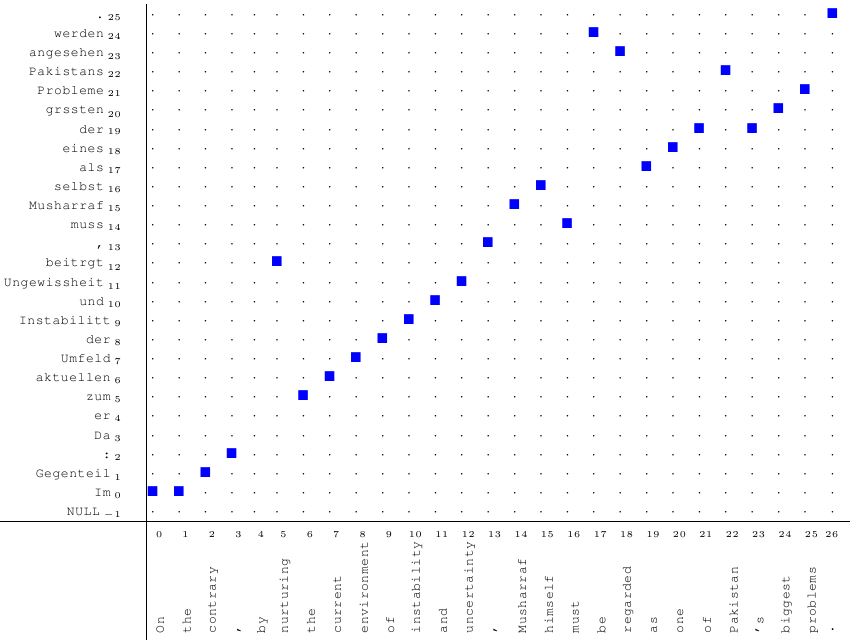}
\hspace{4mm}
\includegraphics[width=.40\textwidth]{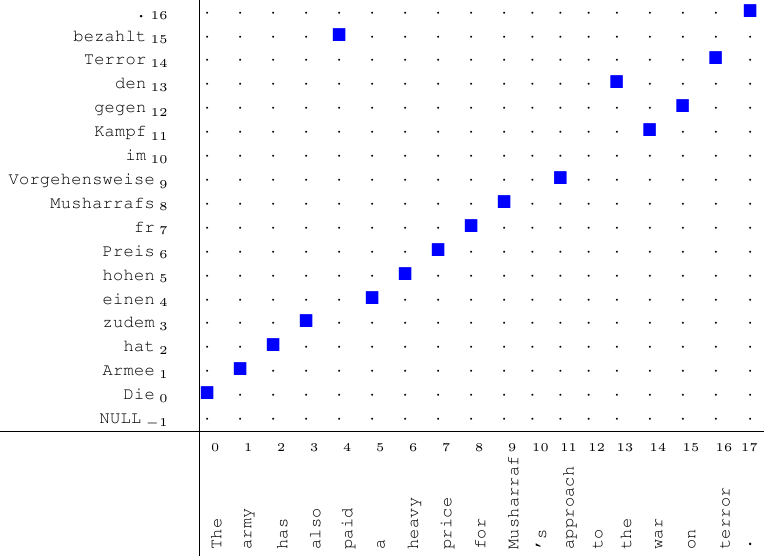}

\vspace{4mm}
English and Arabic:
\vspace{4mm}

\includegraphics[width=.50\textwidth]{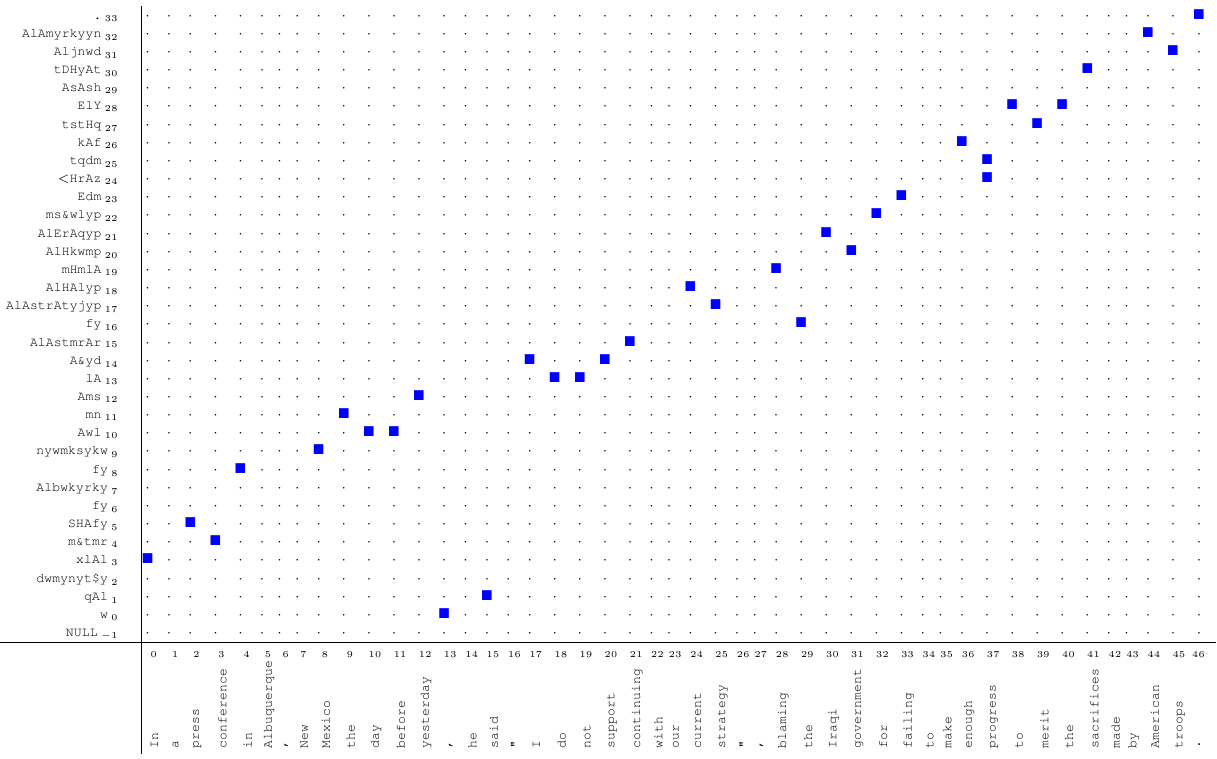}
\hspace{4mm}
\includegraphics[width=.45\textwidth]{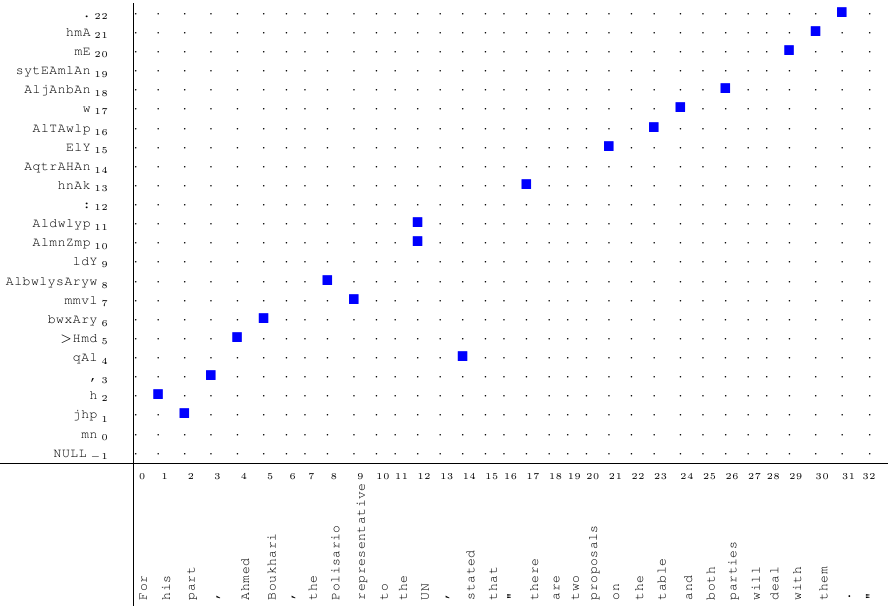}

\vspace{4mm}
English and Turkish:
\vspace{4mm}

\includegraphics[width=.52\textwidth]{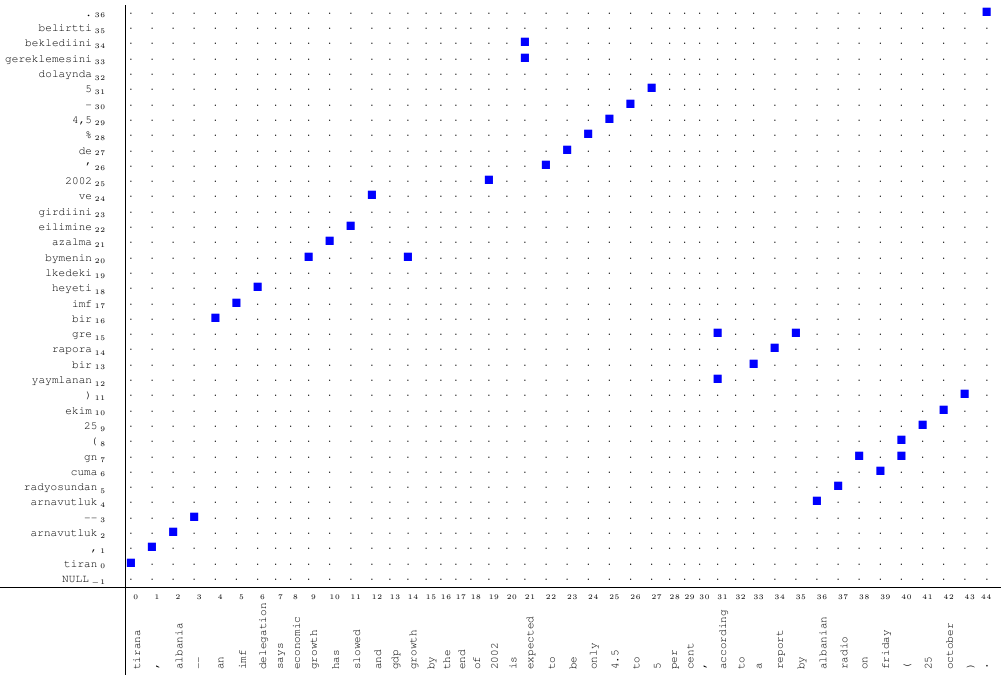}
\hspace{4mm}
\includegraphics[width=.43\textwidth]{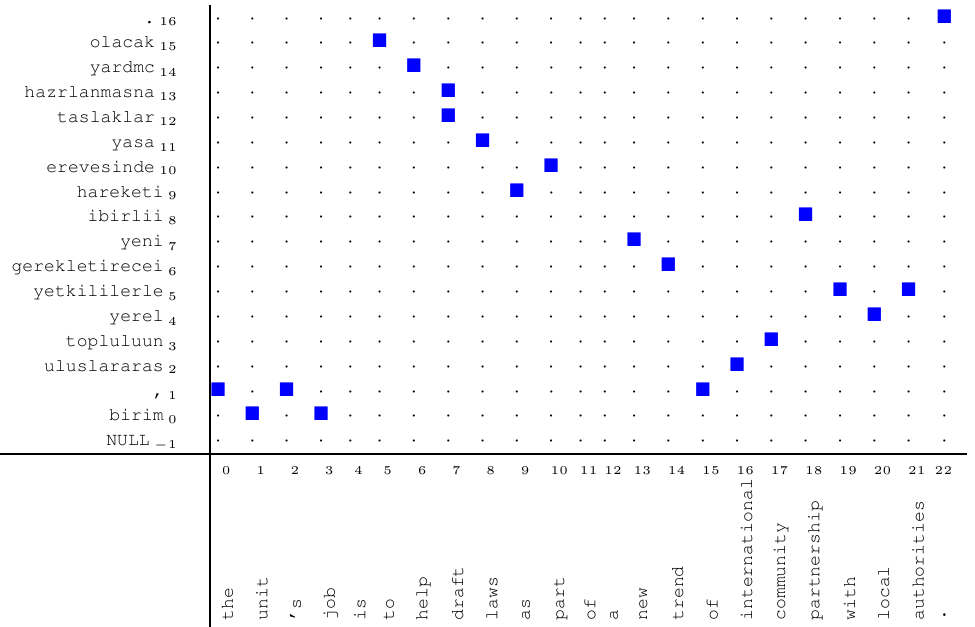}

\vspace{5mm}

\caption{\label{fig:align-matrices} Word-alignment matrices of sentence pairs taken from three parallel news corpora:
the \textsc{nist-mt-08} Arabic-English evaluation benchmark,
the \textsc{wmt-10} German-English training corpus, and
the Turkish-English South European Times corpus (Tyers and Alperen 2010). 
English is always on the \textit{x} axis.}
\end{center} 
\end{figure}


Figure~\ref{fig:align-matrices} illustrates 
the reordering characteristics of three language pairs 
by means of sentence examples that were automatically word-aligned with GIZA++ \cite{Och:03} (intersection of direct and inverse alignments).
On the first row, we see two English-German sentence pairs: 
in both cases, most of the points lie close to the diagonal representing an overall monotonic translation, 
whereas few isolated points denote the very long-range reordering of verbs.
%
Similarly, in the two English-Arabic sentence pairs, we mostly observe local reorderings, with the exception of few isolated points corresponding to the Arabic clause-initial verbs.
Finally, the two Turkish-English examples display global reordering, due to the high number of clause-level order differences.

\nocite{Tyers:10} 

\gap

Where possible, it is interesting to relate our analysis with previously published measures of reordering based on parallel data.
To our knowledge, the most comprehensive results of this kind are reported by \namecite{Birch:11:thesis},
who formulates reordering as a binary process occurring between two blocks that are adjacent in the source (cf. ITG constraints in Section~\ref{sec:sota-PSMT}).
Here, the general amount of reordering in a language pair is estimated by the RQuantity,
defined as the sum of the spans of all the reordered blocks on the target side, normalized by the length of the target sentence and averaged over a corpus.
Based on the Europarl corpus \cite{Koehn:02b} and automatic word alignments,
\namecite{Birch:11:thesis} reports average RQuantity values of 0.586/0.608 in English-to-German/German-to-English, 
versus only 0.402/0.395 in English-to-French/French-to-English.
The manually-aligned GALE corpus (LDC2006E93) is instead used to measure the distribution of reordering widths, defined as the sum of the swapped blocks' target spans. Widths are binned into short (2-4 words), medium (5-8), and long (>8).
In Chinese-to-English there are about 0.8/0.9/0.9 short/medium/long reordered blocks per sentence, 
while in Arabic-to-English there are 1.1/0.4/0.2 short/medium/long reordered blocks per sentence.
%
%
These figures align nicely with our classification of phrase- and clause-level differences,
which we have related to longer and shorter-range reordering respectively:
Chinese-to-English (PDiff: 1, CDiff: 3.5) displays much more reordering overall,
while Arabic-to-English (PDiff: 2.5, CDiff: 0.5) has more short reorderings but much less medium and short.

The advantage of using our proposed analysis is that it can be easily extended to other language pairs thanks to the wide coverage of WALS,
whereas data-driven analyses depend on the availability of high-quality word-aligned parallel corpora.


\section{Discussion and conclusions}
\label{sect:conclusion}

We have provided a comprehensive overview of how the word reordering problem is modeled within different string-based and tree-based SMT frameworks, and as a stand-alone task.
To summarize, string-based SMT considers all permutations of the source sentence and relies on separate reordering models to score them.
On the other hand, tree-based SMT 
tightly couples reordering to translation and, during decoding, only or mostly considers word permutations that are licensed by the learnt translation model.
In practice, both approaches apply general heuristic constraints on the maximum reordering width 
to avoid explosion of the search space.

The main weakness of a string-based approach like phrase-based SMT (PSMT) with regard to reordering lies in its coarse definition of the reordering search space.
In this framework, relaxing the distortion limit means dramatically increasing the size of the search space,
making the reordering model's task extremely complex and intensifying the risk of both search and model errors.
As a result, PSMT is generally good at handling local reordering 
but largely fails to capture long-range reordering phenomena. 

As for tree-based SMT, a distinction must be made between methods that extract hierarchical structure directly from parallel data 
and methods that rely on syntactic annotation provided by pre-trained monolingual parsers.
A prominent example of the former is hierarchical phrase-based SMT (HSMT), which models reordering via partially lexicalized translation rules.
While this results in a more principled definition of the reordering search space, HSMT lacks the ability to generalize the learnt reordering patterns from specific lexical clues to whole word or phrase categories.

Finally, reordering may be constrained by syntactic information in the source or target language, or both.
When syntax is used in the source language, reordering is performed by transforming a given parse tree of the input sentence.
When syntax is used in the target language, reordering is allowed only if resulting in a grammatically valid target tree fragment.
Syntactic information 
is adopted by both syntax-based SMT, where the tree is reordered and translated simultaneously,
and by syntactic pre-ordering (or post-ordering) methods, where the tree is transformed before (or after) translation.
The success of these approaches largely depends on the degree of isomorphism of the modeled language pair,
as well as on the parser's performance, which can vary substantially across languages.


\gap

After describing how word reordering is modeled in SMT,
we have questioned why different language pairs appear to need different reordering modeling solutions.
To answer this question, we have outlined the word order profiles of seven widely spoken languages, based on a large body of linguistic knowledge.
Then we have examined their pairwise differences in detail.
Finally, we have used these differences to interpret the empirical findings of previous work 
that evaluated various SMT reordering techniques in those language pairs.

We conclude from our analysis that 
a few linguistic facts can be very useful to predict the reordering characteristics of a language pair
and to select the SMT approach that best suits them.
In particular, string-based PSMT is preferable for language pairs with only constituent-level differences, like French-English, 
as these mostly imply short or medium-range reordering patterns that can be captured by local distortion.
On the other hand, language pairs with many clause-level order differences (\eg Japanese-English, Turkish-English, Chinese-English)
are best handled by tree-based SMT or syntax-based pre-/post-ordering approaches that can handle complex, hierarchical reordering patterns.
While this may seem obvious, we notice that, in the literature, the choice of an optimal SMT framework for a new translation task is often driven by costly empirical trials rather than by linguistic knowledge.
Finally, the pairs with mostly constituent-level differences and only one or few clause-level differences (\eg German-English and Arabic-English) do not fit well into either category.
In sentences without global reordering, HSMT can underperform PSMT, likely due to the much larger search space explored.
At the same time, applying PSMT to such pairs with heuristic reordering constraints 
can lead to systematic errors in the positioning of important elements of the sentence, such as verbs.
Not surprisingly, these language pairs have been the object of a fair amount work aiming at refining the reordering space of both PSMT and HSMT.

Our word order analysis can be easily extended to other language pairs
using the methodology presented in Section~\ref{sect:phenomena}.

\gap

In conclusion, finding a definitive solution to the problem of word reordering implies answering the fundamental research questions of SMT:
Is structure needed to translate? If so, what kind of structure and how should it be used?
A growing part of the research community has converged on a positive answer to the former question, but the latter remains open to date.
While the field keeps evolving around these questions, SMT has already reached the stage of applied language technology.
We hope this survey will provide practical guidelines to the system developers of today and, at the same time, good scientific references to the researchers elaborating the solutions of tomorrow.


\begin{acknowledgments}

We would like to thank 
Alexandra Birch, Marta R. Costa-juss\`{a},
Nadir Durrani, Chris Dyer, 
Adri\`{a} de Gispert, Isao Goto, Spence Green,
Zhongqiang Huang, Maxim Khalilov, Graham Neubig,
Khalil Sima'an, Milo\v{s} Stanojevi\'{c}, Katsuhito Sudoh,
Christoph Tillmann, Taro Watanabe
and Richard Zens, as well as the anonymous reviewers,
for providing valuable feedback on an earlier version of this survey.

\end{acknowledgments}

\starttwocolumn



\end{document}